\documentclass[ettersize,journal]{IEEEtran}
\usepackage{amsmath,amsfonts}
\usepackage{algorithmic}
\usepackage{array}
\usepackage[caption=false,font=normalsize,labelfont=sf,textfont=sf]{subfig}
\usepackage{textcomp}
\usepackage{stfloats}
\usepackage{url}
\usepackage{verbatim}
\usepackage{graphicx}
\def\BibTeX{{\rm B\kern-.05em{\sc i\kern-.025em b}\kern-.08em
    T\kern-.1667em\lower.7ex\hbox{E}\kern-.125emX}}
\usepackage{balance}
\usepackage{multicol}

\begin{document}
\title{Material Fingerprinting: Identifying and Predicting Perceptual Attributes of Material Appearance}
\author{Ji\v r\'\i~Filip, \and
Filip D\v echt\v erenko, \and
Filipp Schmidt, \and 
Ji\v r\'\i~Lukavsk\' y,\\ \and
Veronika Vil\'\i movsk\'a, \and
Jan Kotera and \and
Roland W. Fleming
\thanks{This research has been supported by the Czech Science Foundation grant GA22-17529S.
This work was also partially funded by the Deutsche Forschungsgemeinschaft (DFG, German Research Foundation) - project number 222641018-SFB/TRR 135 TP C1, by the European Research Council (ERC) Consolidator Award `SHAPE' (ERC-CoG-2015-682859), by the European Research Council (ERC) Advanced Award `STUFF' (ERC-ADG-2022-101098225) and by the Research Cluster ``The Adaptive Mind'' funded by the Hessian Ministry for Higher Education, Research, Science and the Arts.\\
J. Filip, V. Vil\'\i movsk\'a, and J. Kotera are with The Czech Academy of Sciences, Institute of Information Theory and Automation, Prague, Czech Republic
(e-mail: \{filip,vilimovska,kotera\}@utia.cas.cz),
F. D\v echt\v erenko and J. Lukavsk\' y are with The Czech Academy of Sciences, Institute of Psychology, Prague, Czech Republic 
(e-mail: \{dechterenko,lukavsky\}@praha.psu.cas.cz),
R. Fleming and F. Schmidt are with Experimental Psychology, Justus Liebig University of Giessen Germany and Centre for Mind, Brain and Behaviour, Universities of Marburg, Giessen and Darmstadt (e-mail: \{roland.w.fleming,filipp.schmidt\}@psychol.uni-giessen.de)
}
}

\markboth{Material Fingerprinting: Identifying and Predicting Perceptual Attributes of Material Appearance}%
{J. Filip et al.}

\maketitle

\begin{abstract}
The world is abundant with diverse materials, each possessing unique surface appearances that play a crucial role in our daily perception and understanding of their properties. Despite advancements in technology enabling the capture and realistic reproduction of material appearances for visualization and quality control, the interoperability of material property information across various measurement representations and software platforms remains a complex challenge. A key to overcoming this challenge lies in the automatic identification of materials’ perceptual features, enabling intuitive differentiation of properties stored in disparate material data representations. We reasoned that for many practical purposes, a compact representation of the perceptual appearance is more useful than an exhaustive physical description.This paper introduces a novel approach to material identification by encoding perceptual features obtained from dynamic visual stimuli. We conducted a psychophysical experiment to select and validate 16 particularly significant perceptual attributes obtained from videos of 347 materials. We then gathered attribute ratings from over twenty participants for each material, creating a 'material fingerprint' that encodes the unique perceptual properties of each material.
Finally, we trained a multi-layer perceptron model to predict the relationship between statistical and deep learning image features and their corresponding perceptual properties.
We demonstrate the model's performance in material retrieval and filtering according to individual attributes.
This model represents a significant step towards simplifying the sharing and understanding of material properties in diverse digital environments regardless of their digital representation, enhancing both the accuracy and efficiency of material identification.
\end{abstract}


\section{Introduction}
\label{sec:intro}

The digital representation of materials plays a pivotal role in numerous applications, ranging from virtual reality and industrial design to quality control. However, accurately predicting the perceived attributes of these materials from a human vision perspective remains a significant challenge in contemporary research. This is also the case for predicting human judgments automatically based only on limited image information (e.g., a single image or video). This difficulty in mapping physical measurements to perceptual attributes results both from the variety and complexity in material appearances as well as the rich space of human perceptual inferences. Moreover, to be useful, any representation of the  visual properties should be intuitive enough to provide the user with a clear definition of a set of material attributes applicable to material comparison or retrieval.  We call such a representation the ‘visual fingerprint’ of a material.

This paper seeks to achieve this representation through a combination of psychophysical studies, image processing and machine learning methods.
First, we identified some of the most critical appearance attributes of a diverse set of real-world materials including, but not limited to, fabric, leather, wood, plastic, metal, and paper, and use these to characterize the space of appearances. We selected the samples to cover a broad spectrum of textures, colors, and reflective properties, and imaged them to produce standardized video sequences, to provide a comprehensive overview of material appearances typically encountered in both everyday life and specialized industries.

\begin{figure*}[!ht]
\includegraphics[width=\textwidth]{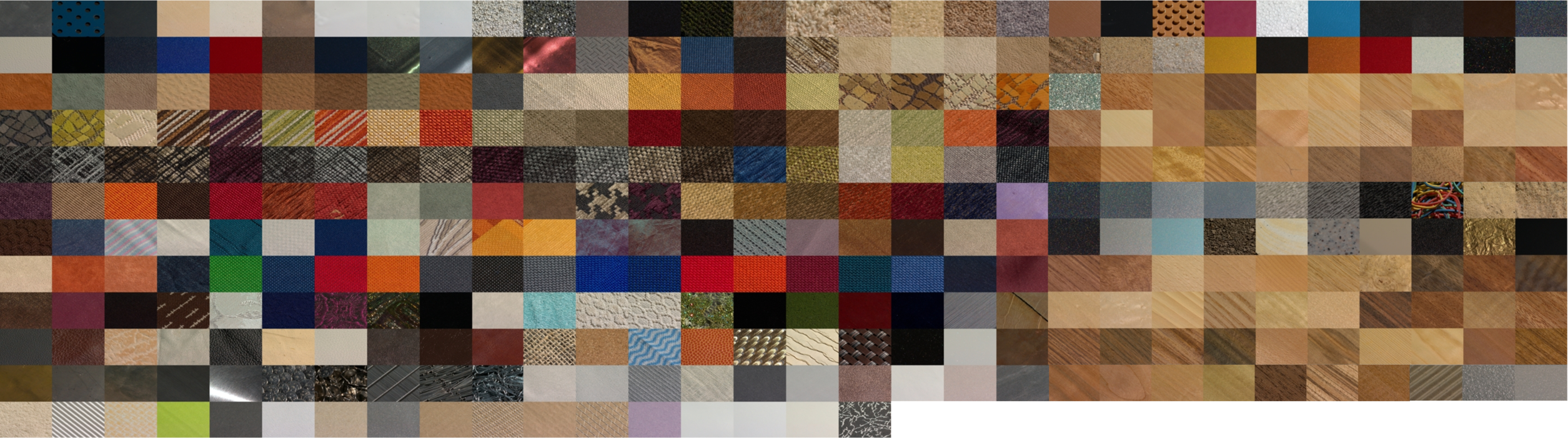}
\caption{\label{fig:mats}A mosaic of all 347 materials in the study, depicted by a single, non-specular frame (\#30 out of 60) from the video sequences.}
\end{figure*}

Instead of using only static images, we opted for captured videos showing the genuine material appearance of flat specimens under different viewing conditions \cite{filip24perceptual}. These dynamic material appearance data allowed us to obtain reliable identification of 16 key appearance attributes. For these attributes, we collected ratings for 347 materials spanning a wide range of categories as shown in Fig.~\ref{fig:mats}, and developed a predictive model of the ratings based on CLIP image features \cite{radford21learning} combined with a multi-layer perceptron model trained on the rating data. With the rise of deep learning, our fingerprint provides an additional interpretive layer that clarifies the relationship between image data and deep learning features for the specific use case of material representation and understanding.

Our findings reveal that combining widely-used and easily-computed image features with a straightforward machine learning model can predict human judgments of appearance and similarity (visual fingerprints) for a diverse range of materials. This opens new possibilities for perceptually-based intuitive and interoperable representations of materials for all applications where material appearance is important. Most notably, our results demonstrate that visual fingerprints can be effectively deduced from just two images of the material, which may have implications for the efficiency and practicality of material analysis in digital applications.

The primary contributions of our paper are:
\begin{itemize}
\item We have created an extensive public dataset of 347 dynamic material samples, significantly enriching available resources for material studies.
\item We have identified 16 crucial perceptual attributes of these materials, providing a foundational understanding of material perception across diverse appearances.
\item We have amassed a substantial dataset by obtaining over 110,000 ratings from human observers for the 16 attributes across all material samples, offering a rich basis for further analysis.
\item Utilizing a multi-layer perceptron model, we have successfully developed a reliable method for predicting rating values. This was achieved by learning a nonlinear mapping from CLIP features, demonstrating the efficacy of machine learning in predicting human perception.
\end{itemize}

\section{Related Work}
\label{sec:related}

Our work relates to human visual perception of material appearance as influenced by illumination and viewing conditions. Specifically, it involves the identification of visual appearance attributes, their approximation through computational features, and the variations of these attributes across different material categories.

\noindent
\textbf{Perception of texture attributes --}
Since the beginning of image texture research, scientists have tried to establish a connection between the perceptual texture space and computational statistics. Tamura et al. \cite{tamura78textural} suggested six basic computational texture properties and evaluated their performance in a perceptual experiment on 56 gray-scale textures from Brodatz's catalogue \cite{brodatz66}. Rao and Lohse \cite{ravishankar96towards} used hierarchical cluster analysis, non-parametric multi-dimensional scaling (MDS), classification and further analyses to group the same textures and identify a perceptual texture space. They concluded that this space can be represented by three dimensions describing repetitiveness, contrast/directionality, and coarseness/complexity. These findings were confirmed by Mojsilovic et al. \cite{mojsilovic00vocabulary} in an experiment with human observers, where they obtained a pattern vocabulary governed by grammar rules, which extended the scope to color textures.

Later, researchers aimed to link perceptual texture spaces to spaces of computational texture features. Malik and Perona \cite{malik90preattentive} used a vocabulary-based method and a model of human preattentive texture perception, based on low-level human vision, to predict the perception of different textures. Vanrell and Vitria \cite{vanrell97afour} suggested a texton-based four-dimensional texture space with perceptual texton attributes along each dimension. In follow-up work, they \cite{vanrell97multidimensional} performed dimensionality reduction of the texton representations to create a low-dimensional behavioral texture space where distances between points represent texture distances. Long and Leow \cite{long02hybrid} presented an approach reducing Gabor features, represented by a convolutional neural network, to a four-dimensional texture space.

Researchers have also studied the human perception of specific texture attributes. Padilla et al. \cite{padilla08perceived} developed a model of perceived roughness in synthetic surfaces. Pont et al. \cite{pont07texture} found that observers use texture as a cue for relief depth and that surface roughness can be exploited to increase the realism of standard 2D texture mapping. Ho et al. \cite{ho07conjoint} found that roughness perception is correlated with texture contrast.

\noindent
\textbf{Perception of textureless material reflectance --} 
Many studies represent material appearance locally or globally by means of the bidirectional reflectance distribution function (BRDF) \cite{nicodemus77geometrical} and its parametric models \cite{guarnera16brdf}. Again, scientists have tried to map the model's parameters to the human perception of materials. Pellacini et al. \cite{pellacini00toward} linked parameters of the Ward BRDF model to the perceived gloss of homogeneous surfaces by creating a two-dimensional perceptually uniform space. Similarly, Westlund and Meyer \cite{westlund01applying} extended the Phong, Ward, and Cook-Torrance BRDF models to a parameterization that allows a more perceptually uniform manipulation of the models' parameters. Matusik et al. \cite{matusik03data} captured and psychophysically evaluated large sets of BRDF samples, forming the MERL BRDF database, and showed that there are consistent transitions of the perceived properties between different BRDFs. They analyzed whether the samples possess any of the 16 predefined perceptual attributes and used the participants' characterizations to build a model in both linear and non-linear embedding spaces. This manifold is then used for editing/mixing between the measured BRDFs.
Wills et al. \cite{wills09toward} analyzed 55 BRDFs from the MERL BRDF database in an extensive experiment with 75 participants assessing glossiness and used MDS to construct a low-dimensional embedding allowing perceptual parameterization of an analytical BRDF model. Te Pas and Pont \cite{pas04comparison} showed that changes in surface BRDF and illumination are often confounded, but adding complex illumination or 3D texture improves visual matching. The dependency of the perception of a light source direction on surface BRDF and its 3D shape was shown in \cite{pas05estimations}. Ramanarayanan et al. \cite{ramanarayanan07visual}, in a series of psychophysical experiments, presented the concept of visual equivalence by characterizing conditions under which warping and blurring of the illumination maps and warping of the object's surface yield synthetic images that are visually equivalent to the reference solutions.

Later, Serrano et al. \cite{serrano18intuitive} psychophysically analyzed isotropic BRDFs from the MERL database \cite{matusik03data} to identify smooth and intuitive material appearance transitions between different visual attributes. Sawayama et al. \cite{sawayama19visual} created a dataset of synthetic images with variable illumination and geometries and conducted perceptual experiments discriminating materials on one of six dimensions (gloss contrast, gloss distinctness of image, translucent vs. opaque, metal vs. plastic, metal vs. glass, and glossy vs. painted). They concluded that material discrimination depended on the illuminations and geometries and that the ability to discriminate the spatial consistency of specular highlights in glossiness perception showed larger individual differences than in other tasks. They also demonstrated that the parameters of higher-order color texture statistics can partially explain task performance.

Lagunas et al. \cite{lagunas19similarity} gathered observers' similarity judgments of synthetic images depicting objects with varying materials, shapes, and illuminations, and trained a deep learning model to measure the similarity in appearance between different materials, which correlates with human similarity judgments and outperforms existing metrics. In follow-up work, Serrano et al. \cite{serrano21effect} collected a large-scale dataset of perceptual ratings of predefined appearance attributes (glossiness, contrast of reflections, sharpness of reflections, metallicness, lightness) for combinations of material, shape, and illumination, to analyze the effects of illumination and geometry on material perception. The collected dataset was used to train a deep learning architecture predicting perceptual attributes that correlate with human judgments. Finally, Subias and Lagunas \cite{subias2023wild} proposed a single-image appearance editing framework based on generative models that allows intuitive modification of the material appearance of an object by increasing or decreasing high-level perceptual attributes describing such appearance (e.g., plastic, rubber, metallic, glossy, bright, rough, and the strength and sharpness of reflections).

\noindent
\textbf{Perception of digital representations of textured materials --}
Extending perceptual analysis to general textured materials introduces a variety of non-local effects such as inter-reflections, masking, shadowing, and subsurface scattering. These effects add visual realism but make experimental work substantially more complex. Jarabo et al. \cite{jarabo14effects} conducted perceptual experiments to investigate the visual equivalence \cite{ramanarayanan07visual} of rendered images for different levels of bidirectional texture function (BTF) \cite{dana99reflectance} filtering, and found that blur in the spatial domain is less tolerable than in its angular counterpart. They analyzed whether BTF datasets possess properties such as \emph{high-contrast}, \emph{granular}, and \emph{hard}, among others.
Deschaintre et al. \cite{deschaintre23visual} introduced a novel dataset that links free-text descriptions to various fabric materials. The dataset comprises 15,000 natural language descriptions associated with 3,000 corresponding images of fabric materials, obtained from observers in the form of free-text descriptions of fabric appearance. By analyzing the data, the authors identified a compact lexicon, a set of attributes, and key structures that emerge from the descriptions explaining how people describe fabrics. Such annotated image data were used to train a large language model to create a meaningful latent space for fabric appearance, appropriate for material retrieval or automatic image annotation.
Finally, Filip et al. \cite{filip24perceptual} analyzed the perceptual dimensions of 30 wood materials in the form of video stimuli by means of a combination of similarity and rating studies and compared them to basic image statistics. In follow-up work \cite{filip23characterization}, they performed a rating study and linked the rating data with computational statistics, demonstrating the extent to which computational statistics can be used to characterize visual properties on an additional test dataset.

\noindent
\textbf{Visual perception across material categories --} 
Recent studies have also analyzed material appearance perception as a function of material type. Fleming et al. \cite{fleming13perceptual} presented an extensive analysis of human perception of materials. In their first experiment, participants judged nine perceptual qualities, while in their second experiment, observers assigned 42 adjectives describing material qualities to six classes of materials. The authors found that the distributions of material classes in the visual and semantic domains are similar and concluded that perceptual qualities are systematically related to material class membership.
In a follow-up study, Tanaka et al. \cite{tanaka15investigating} analyzed human ratings of the same perceptual qualities as a function of visual information degradation. The authors concluded that general perceptual quality decreased with image-based reproduction, perceptual qualities of images decreased when using their gray-scale variant, and perceptual qualities of \emph{hardness} and \emph{coldness} increased when image resolution was reduced.
As an alternative to assigning intuitive attributes to materials, some researchers used material image features encoding material-specific characteristics. Schwartz and Nishino \cite{schwartz13visual} introduced the concept of visual material traits, encoding the appearance of characteristic material properties by means of convolutional features of trained patches, to  identify local material properties in material recognition tasks. In follow-up work, researchers discovered a space of locally-recognizable material attributes from perceptual material distances by training classifiers to reproduce this space from image patches \cite{sharan13recognizing}, \cite{schwartz13automatically}. Later, Schwartz and Nishino \cite{schwartz19recognizing} avoided a fixed set of attributes by proposing a method for deriving material attribute annotations based on probing human visual perception of materials by asking simple yes/no questions comparing pairs of small image patches. This method can be integrated into the end-to-end learning of a material classification CNN to simultaneously recognize materials and discover their visual attributes.

Various aspects of human perception of material appearance have been extensively studied in the past. What sets our study apart from previous work is that (1) our prediction of interpretable appearance attributes is derived from user studies rather than from non-interpretable visual features, and (2) we use videos of a diverse range of real-world materials, showing a continuous loop of a rotating samples to provide a much richer visual impression than commonly used static or synthetic stimuli.
In this work, we build on our previous research on attribute identification \cite{filip24comprehensive} and extend it to the automatic prediction of these attributes from image data.

\section{Method Overview}
\label{sec:data}

\begin{figure*}[!ht]
\begin{center}
\includegraphics[width=0.9\textwidth]{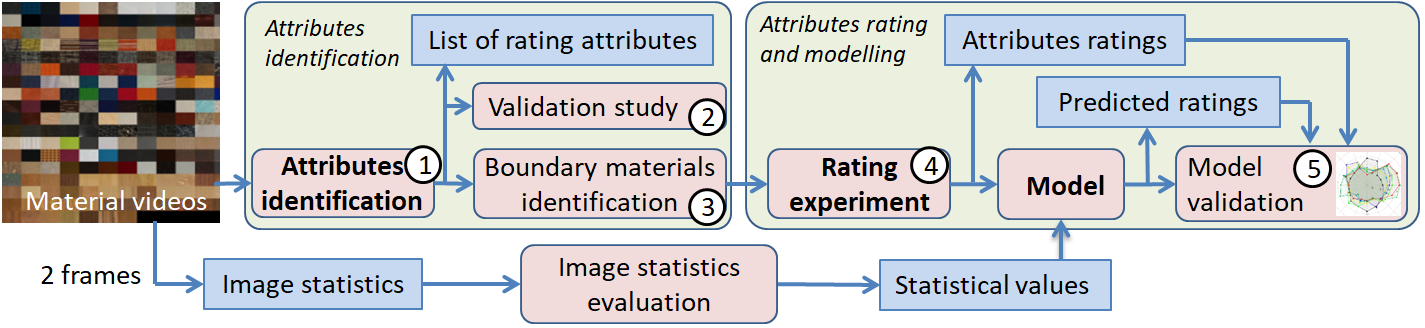}
\end{center}
\caption{\label{fig:scheme}Material characterization by five psychophysical studies to obtain a \emph{visual fingerprint} for 347 material videos and prediction of those attributes with an image-computable machine learning model. Studies and modeling are marked red, resulting data is marked blue.
}
\end{figure*}

The entire process of material characterization using a limited set of intuitive features is depicted in Fig~\ref{fig:scheme}. First, we captured material videos. Then, we identified a set of intuitive perceptual attributes (Studies 1 and 2) and their anchor scales (Study 3). In the following, we obtained attribute ratings for all materials (Study 4), yielding a ‘visual fingerprint’ for each video. To generalize beyond our stimuli, we predicted individual fingerprints with an image-computable model, and evaluated its performance in a validation study (Study 5) and in a real-world task using a new set of material photographs.

\section{Material Data Capturing}
\label{sec:data}

\subsection{Materials Selection}
We collected 347 physical material samples, with a focus on capturing a broad variety of visual appearances but also the most common material categories.
Although the potential pool of textured materials is vast, we prioritized those for which additional appearance measurements are available, such as benchmark materials from the UTIA BRDF database \cite{filip14template} and the MAM 2014 collection \cite{rushmeier14MAM}.
For many material categories with spatially homogeneous appearances, such as metal, plastic, and paper, individual materials can be relatively accurately represented using parametric reflectance models \cite{guarnera16brdf}. Here we focused on more visually complex materials exhibiting non-local physical effects like shadowing, masking, or subsurface scattering, so that the majority of materials in our selection were woods and fabrics that vary strongly in appearance due to different fiber types \cite{filip24perceptual} and thread weaving patterns \cite{filip14template}. Also for the remaining categories, we focused mainly on material samples with non-homogeneous structures.
Our final selection consisted of 347 physical samples from the material categories of fabric (157), wood (67), coating (30), paper (23), plastic (17), metal (14), leather (11), and others (28) (see Fig.~\ref{fig:mats})

\subsection{Dynamic Stimuli Capturing}
To take into account the role of real-world illumination and the interactions between lighting and object geometry in the estimation of material properties \cite{fleming03realworld}, \cite{fleming14visual}, we created dynamic stimuli. Specifically, for each material sample, we produced a video sequence showcasing the material's non-specular and specular characteristics by a slow rotation. These sequences featured close-up views of approximately 42$\times$42 mm areas of the samples, captured using the UTIA goniometer \cite{filip13brdf}. In line with industry standards \cite{mccanny96observation}, we maintained a constant polar angle of 45 degrees for both the camera and the light source, varying only the azimuthal angle of the camera to facilitate more rapid measurements. Each sequence commenced with the light and camera azimuthal angles differing by 90 degrees, followed by a 90-degree camera movement, resulting in a final difference of 180 degrees between the azimuthal angles probing specular reflection of the material. Comprising 60 image frames of resolution 632$\times$412, each 4-second sequence was played in reverse order after completion, creating an 8-second continuous loop that effectively illustrates the dynamic behavior of the ‘rotating’ material sample (cf. video in the supplementary material).

\begin{figure}[!ht]
\includegraphics[width=\columnwidth]{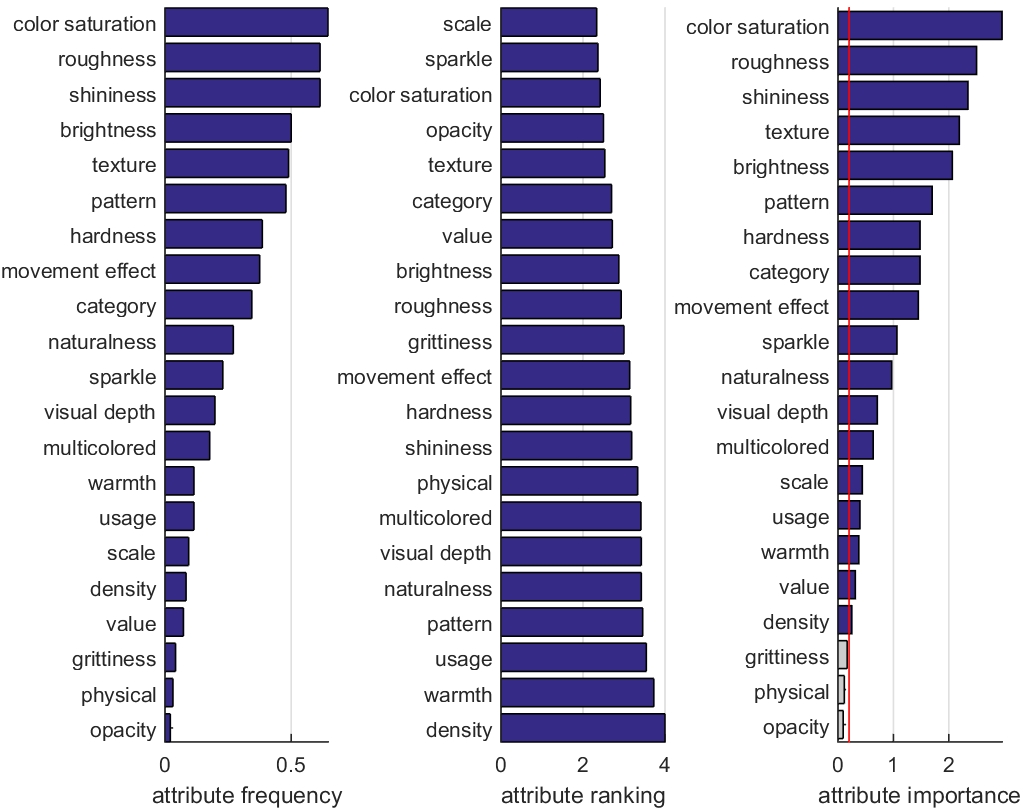}
\caption{\label{fig:attribStats}Attribute statistics obtained from the free naming experiment: attribute frequency (left), ranking (center), and the combined importance of each attribute (right). }
\vspace{-0.5cm}
\end{figure}

\section{Selection of Main Perceptual Attributes}
\label{sec:attribs}

\subsection{Study 1 -- Attributes Identification}

First, we identify the key visual attributes for describing the appearances of material videos in our dataset (also see Fig.~\ref{fig:scheme}).

For the online free naming study, we created three arrangements of 70 material videos each, randomly selected from our full dataset. Participants were asked to type at least five most visually distinguishing features, that they thought distinguishes best between the materials in each arrangement, and rank them in the order of their importance. Each participant responded to all three arrangements in different trials. We collected a total of 451 text responses from 32 participants. By manually grouping synonyms and equivalent terms into clusters, and removing all responses occurring less than twice (eliminating 0.44\% of total responses), we obtained a condensed set of 21 visual material attributes (raw responses for each attribute are provided in the supplementary material).

Fig.\ref{fig:attribStats}shows the frequency $a_p$ of each attribute (left; calculated across participants and trials), as well as the average rank $a_o$ (center), and the combined attribute importance (right), calculated by $a_p \cdot (\max(a_o) - a_o) $. As a result, the participants described the visual appearances of our material videos by using typical optical attributes such as color variability, saturation, roughness, brightness, shininess, texture, and pattern, but also tactile or cognitive attributes like warmth, hardness, naturalness, and attractiveness.
As two of the most frequent terms \emph{texture} and \emph{pattern} are vague without further elaboration we replaced them with the more specific attributes \emph{pattern complexity}, \emph{striped pattern}, and \emph{checkered pattern}. The clusters\emph{gritty}, \emph{physical}, and \emph{opacity} were removed as they were rarely mentioned (less than five responses). We also removed clusters \emph{usage}, \emph{category} describing utilization or class of the material, and \emph{density} was merged with \emph{hardness}.
In total, our clustering included 98.5 \% of all raw naming responses. The resulting final set of 16 perceptual attributes is shown in
 Tab.~\ref{tab:attribs}, together with the boundary (anchor) terms and the instructive questions that were provided to the participants in our following rating study.

\begin{table*}
\renewcommand{\arraystretch}{0.5}
\caption{\label{tab:attribs}A list of 16 perceptual attributes evaluated in the rating study and their description.}
\begin{tabular}{|*{3}{l}|*{1}{p{10.5cm}}|}
\hline
ID & attribute & boundary values & instructive question \\
\hline
1.&\textbf{color vibrancy}      &dull, vibrant     &\small{How richly colored is the material, ranging from monochromatic or neutral-colored materials to vibrantly colored materials?}\\
2.&\textbf{surface roughness}&smooth, rough &\small{How rough is the material, ranging from fine or smooth to coarse or grainy?}\\
3.&\textbf{pattern complexity}&plain, complex &\small{How complex are the patterns on the material, ranging from simple to intricate?}\\
4.&\textbf{striped pattern}     &no stripes, pronounced stripes&\small{To what extent does the material exhibit stripy patterns?}\\
5.&\textbf{checkered pattern}&no checks, pronounced checks&\small{To what extent does the material exhibit checkered patterns?}\\
6.&\textbf{brightness}          &black, white      &\small{How bright is the material, ranging from dim or subdued to bright or luminous?}\\
7.&\textbf{shininess}            &matt, mirror     &\small{How shiny is the material, ranging from dull or non-reflective to highly reflective?}\\
8.&\textbf{sparkle}              &none, sparkling  &\small{To what extent does the material exhibit sparkling and glittery effects?}\\
9.&\textbf{hardness}           &soft, hard         &\small{How hard is the material, ranging from soft or plush to firm or rigid?}\\
10.&\textbf{movement effect}&none, extreme&\small{To what extent does the appearance change due to camera movement?}\\
11.&\textbf{pattern scale}    &fine, large         &\small{How large are the pattern elements, ranging from fine-grained or uniform to large or blotchy patterns?}\\
12.&\textbf{naturalness}      &manmade, natural&\small{How natural is the material, ranging from man-made to natural origin?}\\
13.&\textbf{thickness}         &flat, thick          &\small{How deep is the material structure, ranging from flat or thin to thick?}\\
14.&\textbf{multicolored}      &single, many     &\small{How multicolored is the material, ranging from a single or uniform color to colorful or many colors?}\\
15.&\textbf{value}              &cheap, luxurious&\small{How valuable is the material, ranging from low-cost or cheap to extravagant or luxurious?}\\
16.&\textbf{warmth}           &cold, warm       &\small{How warm is the material to the touch, ranging from cool or cold to pleasant or warm?}\\
\hline
\end{tabular}
\end{table*}

\subsection{Study 2 -- Attributes Validation}

As the clustering of attributes might have been subject to experimenter bias, we performed a second study where we asked six participants to cluster all 451 valid text responses into the obtained 16 attributes (see Tab.~\ref{tab:attribs}). Overall, the inter-rater agreement was notably high (Fleiss' Kappa score of 0.786), and for 198 out of 451 responses (43.9\%), all six raters reached a unanimous decision. For 254 (56.3\%) of responses, at least three raters agreed, and for 396 (87.8\%) responses, at least two raters did. As a consequence, the naming responses of 16 out of 32 participants in the free naming study fit completely into the derived rating attributes, while the divergence for the remaining 16 participants varied in a range between 5.6\% to 26.7\%.

\section{Attributes Rating}
\label{sec:rating}

\subsection{Study 3 -- Boundary Materials Identifications}
To create a representative visual anchor for the rating study, we asked 9 online participants to pick from the three arrangements of 70 material videos those materials with the lowest and the highest value of a specified visual attribute (e.g., Which of the materials displays the highest level of brightness?). Participants completed 96 responses each (3 arrangements$\times$16 attributes$\times$2 boundary anchors). On average, 3.6 (out of 9) participants identified the same video as expressing the lowest value of an attribute, and 2.8 participants for the highest value. After removing multiple occurrences (resulting from a video being picked for more than one attribute), we obtained 25 materials that were used as anchors in the following rating study (Fig.~\ref{fig:ratingStim}).

\subsection{Study 4 -- Rating Experiment}
\begin{figure}[!ht]
\includegraphics[width=\columnwidth]{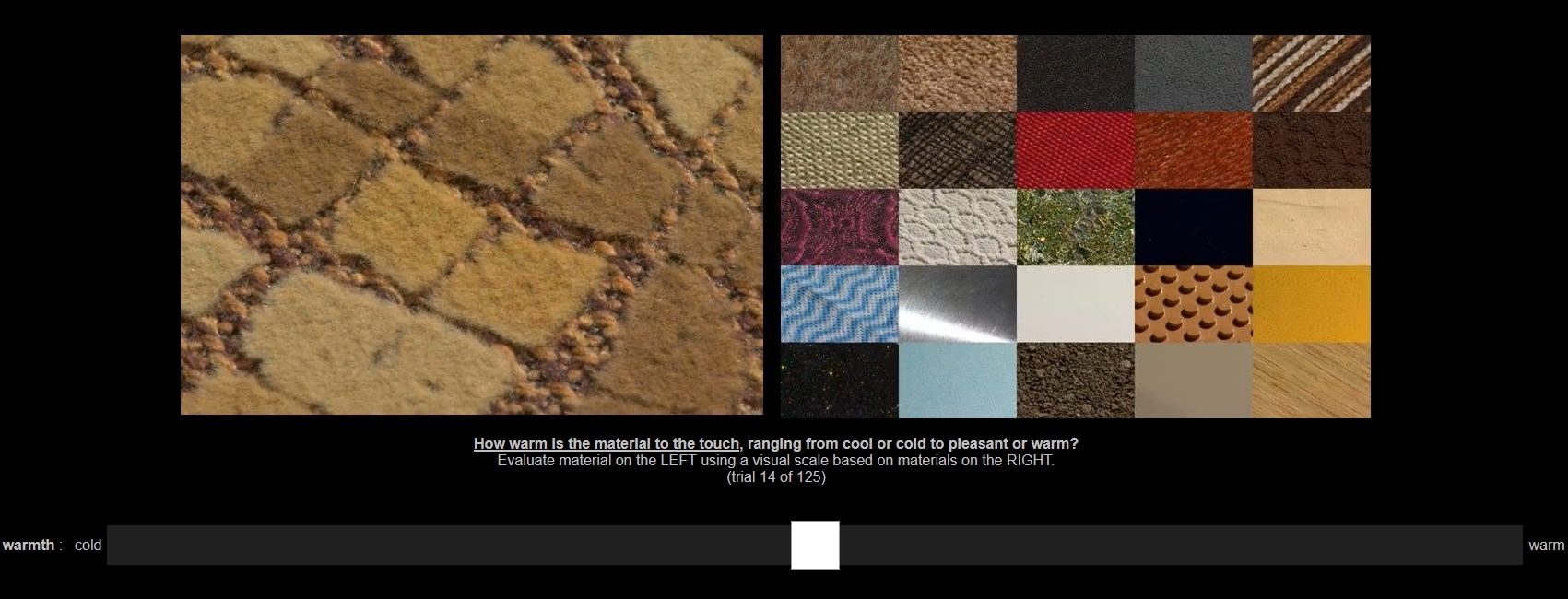}
\caption{\label{fig:ratingStim}An example trial of the rating experiment with the stimulus on the left and the anchor materials on the right (both were videos, depicted here by a single frame).}
\end{figure}

In each trial of the online rating experiment, we showed a material video on the left, together with the arrangement of 25 anchor boundary material videos on the right. For each perceptual attribute, we showed all 347 material videos in random order, and participants responded to the respective question (right column in Tab.~\ref{tab:attribs}) with a slider (Fig.~\ref{fig:ratingStim}). The anchor materials were identical for all tested videos and attributes.

\begin{figure*}[!ht]
\includegraphics[width=\textwidth]{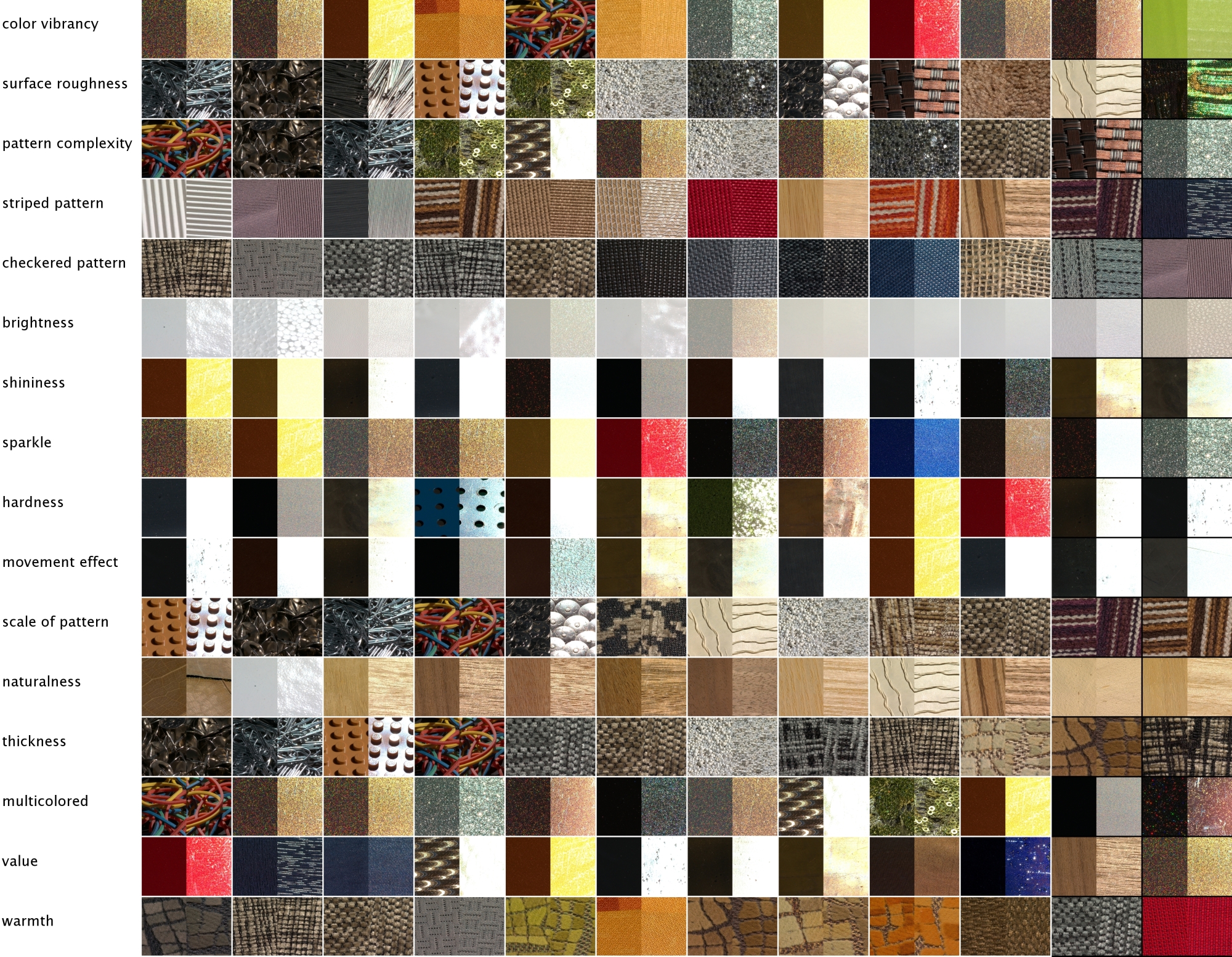}
\caption{\label{fig:matRank}Rank ordering of materials based on their rating values for individual attributes, with the highest value on the left and decreasing values to the right (only showing the 12 highest-rated videos, each depicted by one half of a specular and non-specular frame).}
\end{figure*}

We collected a total of 111,040 ratings (20-24 participants/attribute). After normalizing the data at participant level by Z-scoring, we computed mean rating scores for each material video across all participants. Then, we excluded participants' ratings from the analysis when the correlation between their material ratings for an attribute with the mean material ratings across all participants for that attribute was negative (removing on average 1.5 participants per attribute). The remaining data was averaged across participants to obtain mean ratings for all 16 attributes and 347 materials.
Fig.~\ref{fig:matRank} shows the resulting rank ordering of materials for individual attributes.

\subsection{Material Similarity from Attribute Ratings}

We tested a number of different metrics to evaluate the similarity between two materials in terms of their attribute ratings, e.g. L1, L2, correlation, f-divergences and their combinations. The closest visual matches were obtained for a weighted combination of Pearson correlation ($R$) and L1-norm, where the first term accounts for relative similarity between individual attributes, while the second term compares difference between amplitude of two sets of attributes:
\begin{equation}
d(V_1,V_2) = \alpha R(V_1,V_2) + (1-\alpha) \left[1-\frac{1}{2n}\sum_i |V_1(i)-V_2(i)| \right].
\label{eq:1}
\end{equation}
Where $n$ is length of attribute vector, i.e. 16, $\alpha$ allows the user to prefer either proportional similarity of attribute values ($\alpha=1$) or favor minimal absolute distance ($\alpha=0$) between vectors of ratings. In our experiments we used $\alpha=0.5$.

\subsection{Rating Results}

\begin{figure*}[!ht]
\centering
\includegraphics[width=0.90\textwidth]{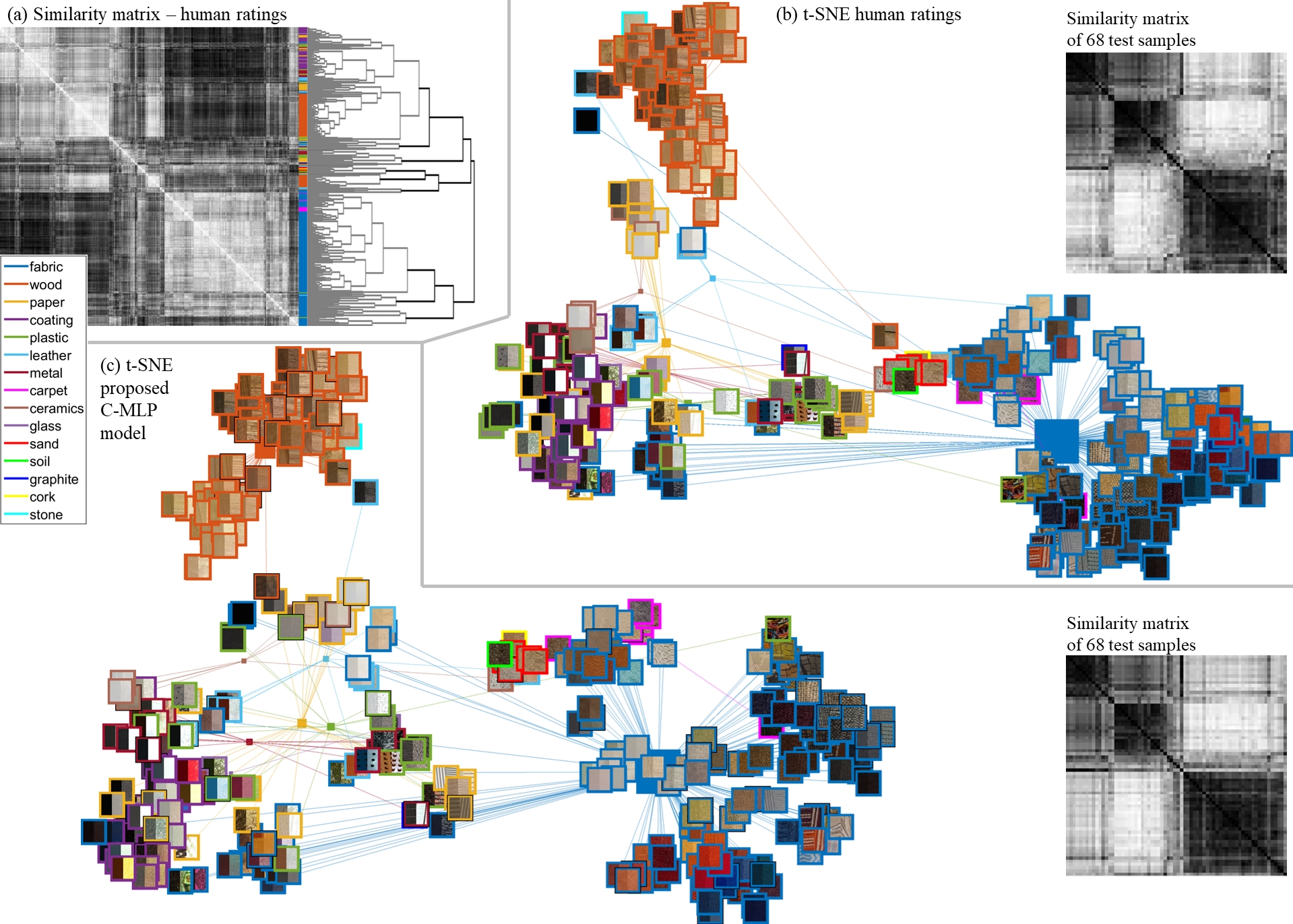}
\caption{\label{fig:main}(a) Similarity matrix between all materials with corresponding dendrogram. (b) Similarity of material samples based on human attribute ratings visualized as two-dimensional embedding using t-SNE. (c) Visualization of the proposed C-MLP model (SM) using t-SNE. Ground truth material categories are color-coded; in the t-SNE plots, category means are plotted as coloured boxes with the size depending on the number of samples. (b) and (c) include the similarity matrices for the 68 test samples obtained from the ratings and the proposed C-MPL model.}
\end{figure*}

We visualize the similarity between material samples using t-distributed stochastic neighbor embedding (t-SNE) \cite{van08visualizing} (Fig.~\ref{fig:main}-b). For the classes of \emph{wood}, \emph{fabric, carpet}, and \emph{coating}, we find coherent clustering according to the ground truth category, while we see considerable overlap for other classes. This follows from a high variability in appearance of samples within those classes (and therefore a high variability in attribute ratings), for example, metal sheets vs. pins from the same material (also see Fig.~\ref{fig:main}-a).

\section{Visual Material Fingerprint}

\begin{figure}[!ht]
(a) material fingerprint
\begin{center}
\includegraphics[width=0.8\columnwidth]{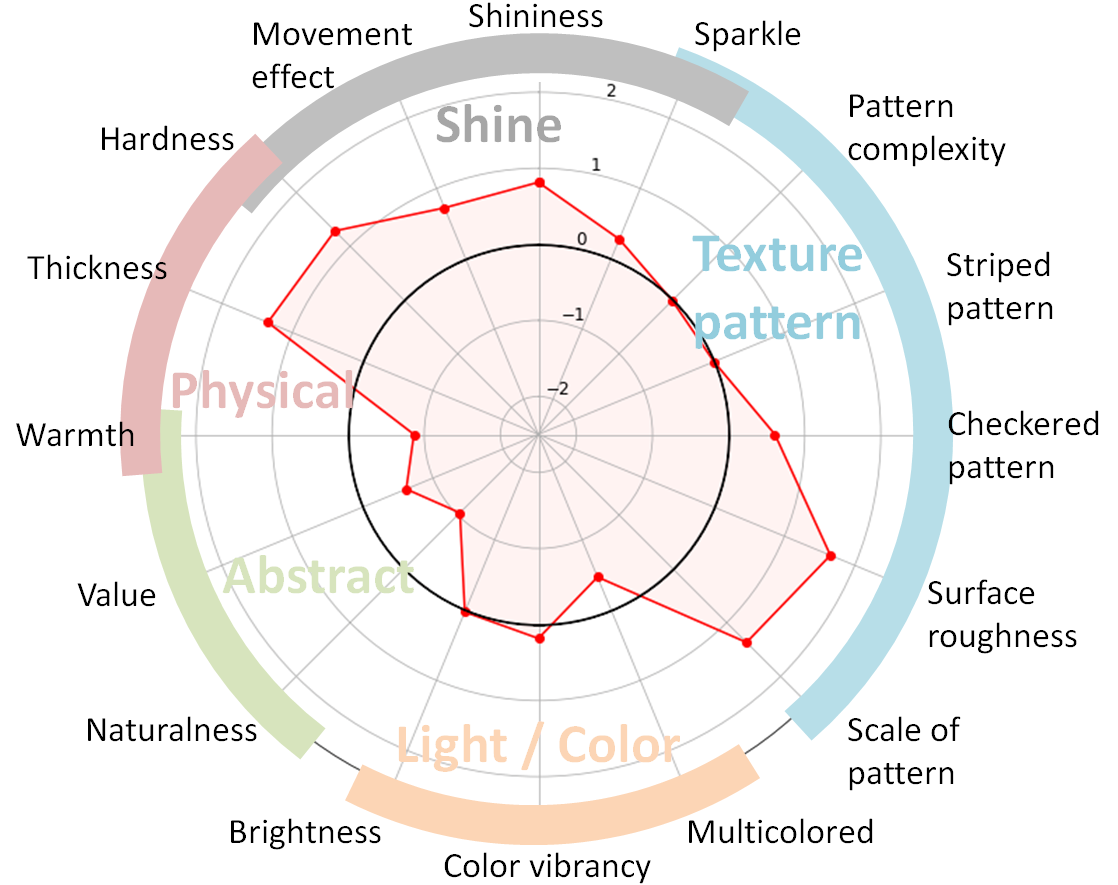}
\end{center}
\vspace{0.1cm}
(b) visual attributes for major material categories:\\
{\small
\begin{tabular}{*{4}{p{1.7cm}}}
\textsf{fabric} & \textsf{wood} & \textsf{paper} & \textsf{coating}\\
\includegraphics[width=0.24\columnwidth]{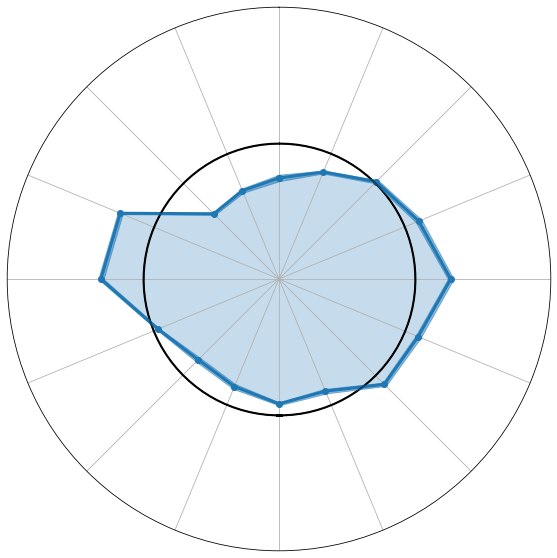}&
\includegraphics[width=0.24\columnwidth]{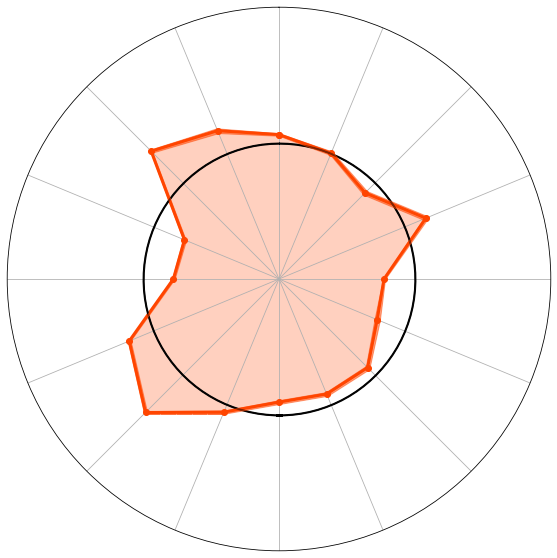}&
\includegraphics[width=0.24\columnwidth]{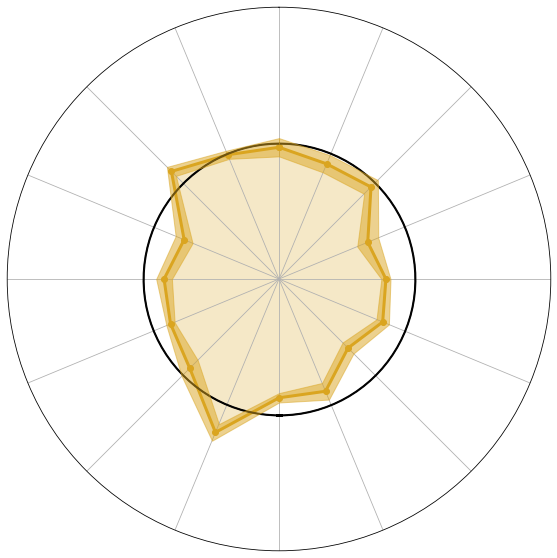}&
\includegraphics[width=0.24\columnwidth]{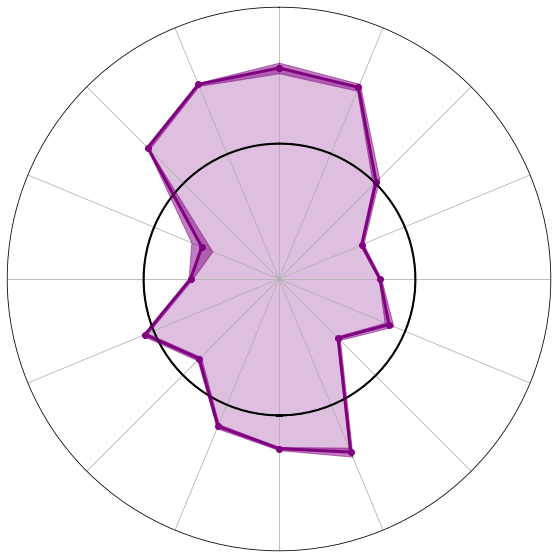}\\
\textsf{plastic} & \textsf{leather} & \textsf{metal} & \textsf{carpet} \\
\includegraphics[width=0.24\columnwidth]{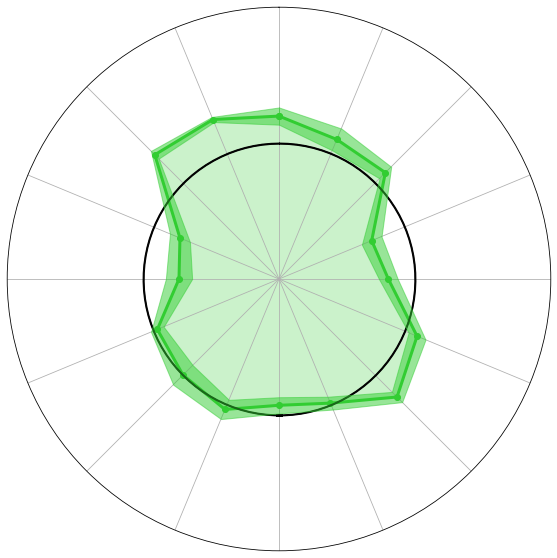}&
\includegraphics[width=0.24\columnwidth]{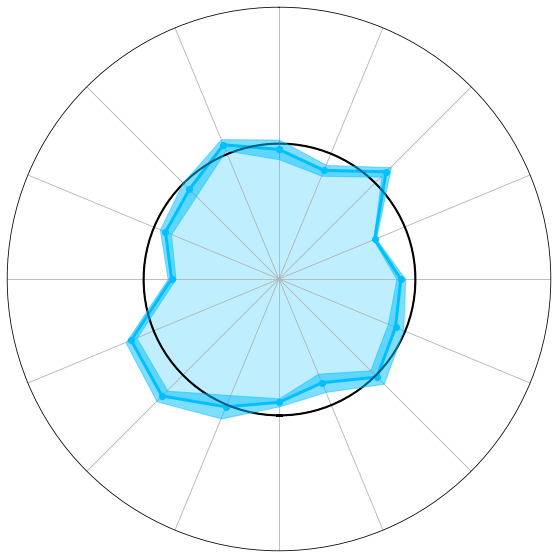}&
\includegraphics[width=0.24\columnwidth]{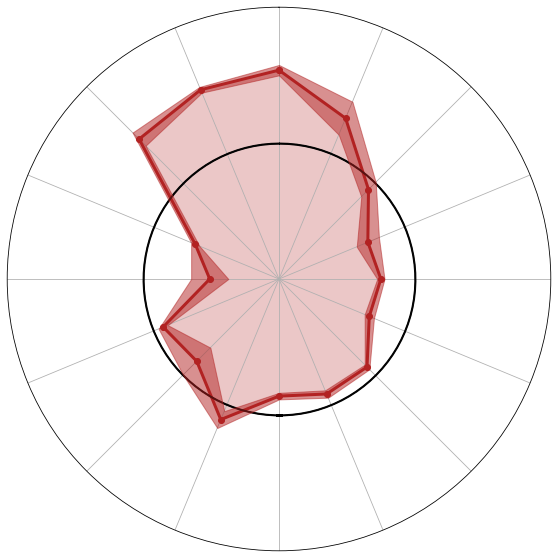}&
\includegraphics[width=0.24\columnwidth]{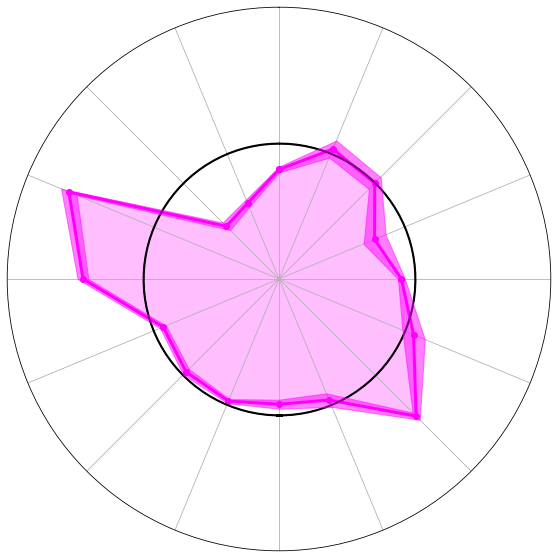}\\
\textsf{ceramics} & \textsf{glass} & \textsf{sand} & \textsf{soil} \\
\includegraphics[width=0.24\columnwidth]{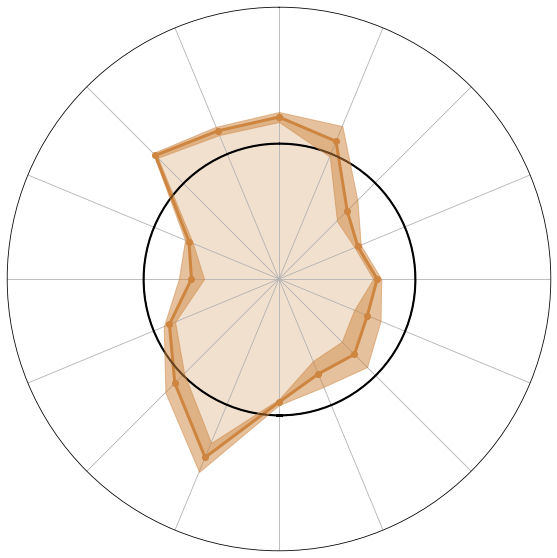}&
\includegraphics[width=0.24\columnwidth]{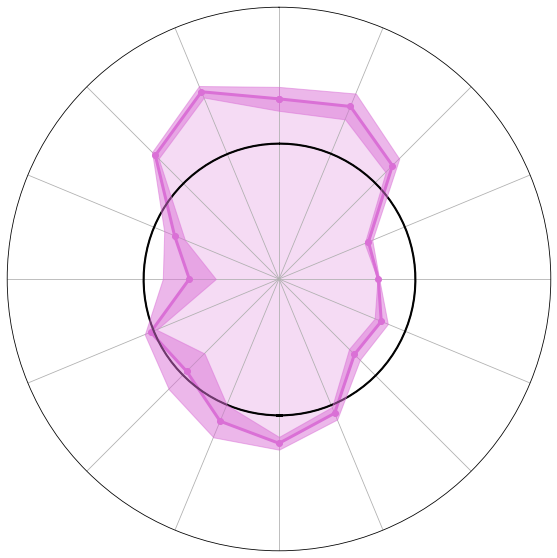}&
\includegraphics[width=0.24\columnwidth]{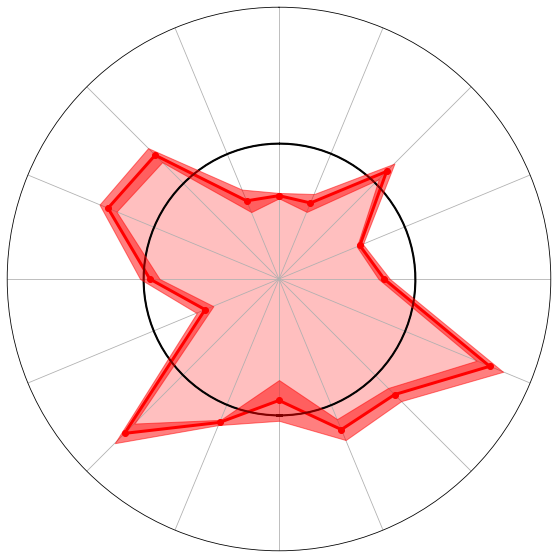}&
\includegraphics[width=0.24\columnwidth]{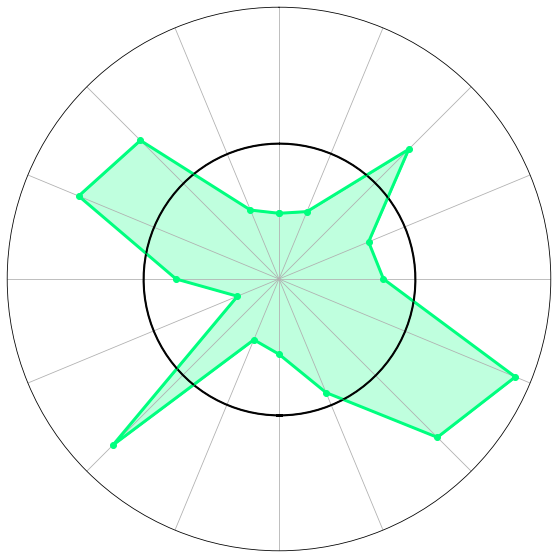}\\
\end{tabular}
}
\caption{\label{fig:ratingSim}(a) An example visual fingerprint of material, and (b) average fingerprints for 12 major material categories, with the thickness of the plot’s contour indicating the standard error in the attribute ratings.}
\vspace{-0.3cm}
\end{figure}

The material attributes determine a unique visual signature of a material's appearance that can be visualized in a polar plot (Fig.~\ref{fig:ratingSim}-a). The azimuthal ordering of attributes is based on their relationships, forming five clusters loosely related to \emph{gloss}, \emph{texture and pattern}, \emph{light and color}, and both \emph{physical} and \emph{abstract} properties. High values are closer to the plot’s boundary, low values are closer to its center. Fig.~\ref{fig:ratingSim}-b displays the median attribute rating values across samples in 12 major material categories from our dataset. This highlights the similarities and differences between the fingerprints of different categories: for example, fabrics and carpets are both characterized as thick and warm, whereas coatings and metals are distinctively not, but rather as hard and shiny.

\section{Human Ratings Prediction }
\label{sec:rel}

In this section, we investigate the possibility of predicting the ‘visual fingerprint’ computationally. Specifically, we propose a method to predict human ratings of visual attributes directly from photographs of material samples. This approach consists of two stages, for which we explored several possibilities. First, we compute a set of intermediate image features, referred to as the ‘image-computable model,’ and then we train a classifier to predict human ratings from these features.

\subsection{Image-computable Representation of Material Images}

We tested several sets of image features computed from the material images. Our experiments revealed that a single image from the video sequences did not provide sufficient information about the material's visual properties, particularly for shiny materials or those exhibiting high dependence on illumination or viewing angle. Therefore, for feature computation, we use two frames from the material videos, corresponding to different observation angles: one for non-specular and one for near-specular behavior. However, the ideal specular angle, where light is reflected directly into the camera, is not suitable for shiny materials, as the corresponding image frame is often saturated. To address this, we offset the azimuthal angle from the ideal specular angle by approximately 6$^\circ$.

\textbf{Deep learning features} – The main proposed model is motivated by the work of Kaniuth et al. \cite{kaniuth24high}, who presented an image-computable model of perceived similarity and demonstrated that the deep features of pretrained vision models perform similarly to humans in assessing visual similarity. Inspired by their results, we used the CLIP model (Contrastive Language-Image Pre-training, \cite{radford21learning}) as a feature extractor.
Specifically, we used the ViT-B/32 backbone with a 224px input image size and a 512-dimensional output feature vector. Each of the two video frames from a material was first resized to 250 DPI, corresponding to 256px on the training set, and then center-cropped to 224px before being processed by CLIP. The output feature vectors were concatenated, resulting in a 1024-dimensional image-computable model vector. In the results, this model is denoted by the prefix C.

\textbf{Statistical features} – In addition to deep features, we also experimented with two different sets of hand-crafted features based on image statistics.

The first set used texture synthesis image statistics introduced by \cite{portilla00parametric}, which are prefixed by T in the results. Since these features enable realistic image synthesis, it is conceivable that they are also suitable descriptors of the visual properties of the material. We used three pyramid levels, three orientations, and a spatial neighborhood of 7x7 pixels for both video frames and color channels separately. After removing features with zero values we ended up with 447 features per channel resulting in a 1341-dimensional feature vector.

Additionally, we extended the compact set of features that we have successfully used in previous work to find mappings between material images and their human ratings \cite{filip23characterization}, prefixed by S in the results. These 14 features (28 for both frames) consist of up to third-order image statistics, energy of frequency bands, measures of directionality, number of dominant colors, and pattern type. More details are provided in the supplementary material.

\subsection{Mapping from Computational Image Features to Human Ratings}

In the second stage, we need to establish the mapping between the intermediate image-computable model and the 16 attributes of human ratings (i.e., the visual fingerprint) of the material. For this, we explored two inference methods:

\textbf{K-nearest neighbor model (2NN)} – For each query material sample, we identified $k$ materials in the training dataset that have the closest (in the $L^2$ sense) image-model features and predict the perceptual ratings by linear interpolation of the ratings of these samples. We obtained the best performance for $k=2$, and we denote this method as 2NN.

\textbf{Multi-layer perceptron model (MLP)} – We trained a small, fully connected neural network to learn the mapping between the intermediate feature representation and the human ratings. We split the dataset of 347 materials into 80\% training and 20\% test samples, resulting in a 279/68 split. The test samples were selected randomly for each material category to ensure that all material categories were present in the test set. We adjusted the model size for each of the three image-computable models to achieve the best performance. Specifically, the layer dimensions used were 1341-256-64-32-16 for the texture synthesis features (T), 28-16-16-16 for the statistical features (S), and 1024-512-512-16 for the CLIP features (C). We used ReLU activations in all layers except the output layer. During training, we applied standard augmentations such as random crop, rotation (except for the statistical features, which are inherently rotation-invariant), small scaling (up to 5\%), and small perturbations of the azimuthal angle (up to 2.5$^\circ$).

\subsection{Results}

We tested different variants of image features (T, S, C) and inference methods (2NN, MLP) but only the best performing are reported in the paper. All computations were done in Python and Pytorch, and implementation details including fingerprint models and inference codes are provided in the supplementary material. The source code is provided on the project website.

\begin{table}
\caption{\label{tab:models}A comparison of the tested models in terms of their correlations with the human similarity matrix $R_{SM}$ and the rating values $R_{A}$, as well as their mean average error.
}
\begin{tabular}{|l|rr|r|}
\hline
  method & $R_{SM}$ & $R_{A}$ & MAE\\
  \hline
  \multicolumn{4}{|l|}{(a) all data (347 samples)}\\
  \hline
  \textbf{T-MLP} texture synthesis   &0.868 & 0.843 & 0.285\\
  \textbf{S-MLP} image statistics      &0.914 & 0.885 & 0.251\\
  \textbf{C-2NN} CLIP model           &0.844 & 0.833 & 0.381\\
  \textbf{C-MLP} CLIP model (proposed) &\textbf{0.971} & \textbf{0.973} &  \textbf{0.117}\\
  \hline
  \multicolumn{4}{|l|}{(b) test samples only (68 samples)}\\
  \hline
  \textbf{T-MLP} texture synthesis       &0.829       & 0.725 & 0.378\\
  \textbf{S-MLP} image statistics     	&0.869        & 0.816 & 0.305\\
  \textbf{C-2NN} CLIP model 		&0.907        & 0.853 & 0.375\\
  \textbf{C-MLP} CLIP model (proposed) &\textbf{0.945} & \textbf{0.914} &  \textbf{0.220}\\
  \hline
\end{tabular}
\end{table}

\textbf{Correlation analyses} --  the second column of Tab.~\ref{tab:models} compares the correlations between the similarity matrices computed from model predictions and human ratings; the third column shows the correlations between the predictions and human ratings (16 attributes$\times$347 materials); while the fourth column shows the mean average errors. The C-MLP model exceeds all other models in predicting the human data.

\begin{figure}[!ht]
\includegraphics[width=\columnwidth]{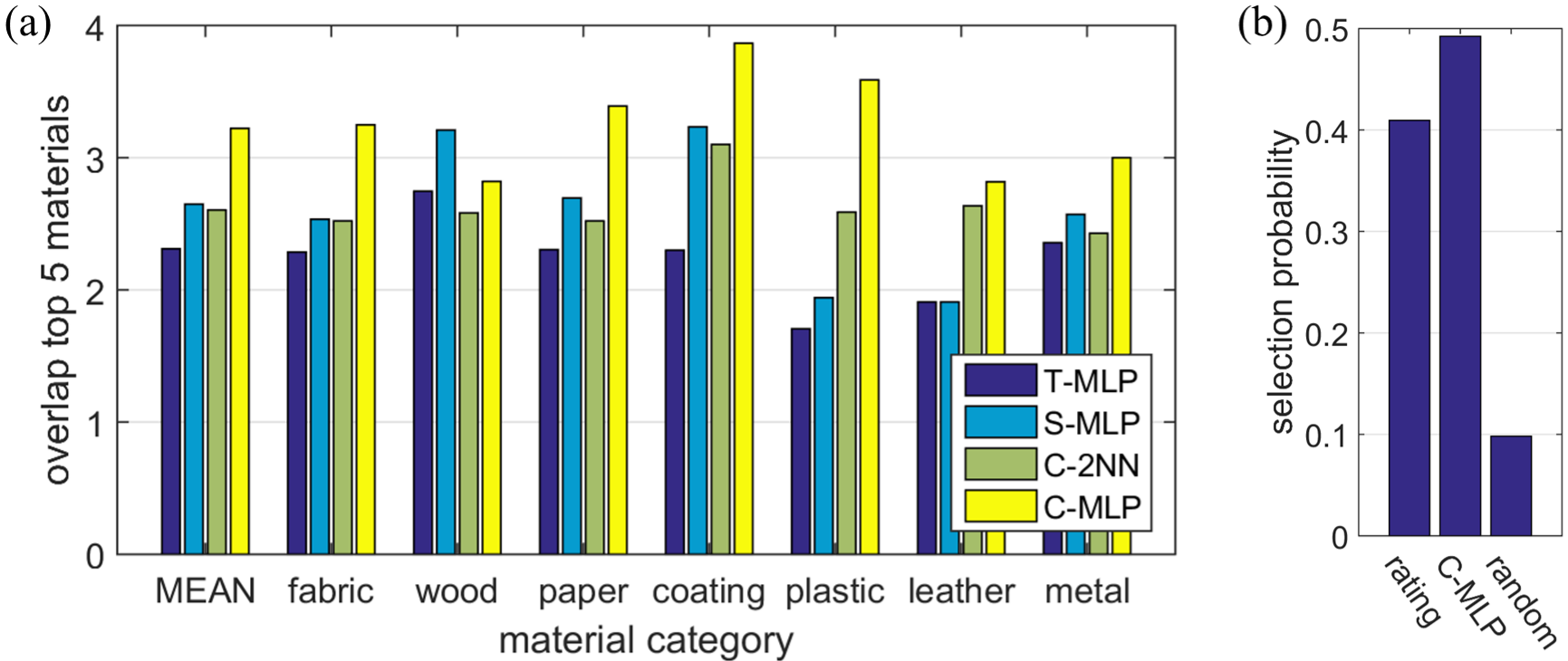}
\caption{\label{fig:top5}(a) Overlap of the top five material images across attributes for model predictions versus human ratings in the test set. (b) Percentage of observers that for a given material sample preferred the closest four material samples selected based on either the human ratings, the proposed C-MLP model, or a random selection.}
\vspace{-0.3cm}
\end{figure}

\textbf{Retrieval overlap} -- 
To further compare all tested models, we also evaluated the overlap between the top five material images in the test set for each attribute based on the model predictions versus the human ratings as shown in Fig.~\ref{fig:search}. As a result, the C-MLP model again clearly outperformed other models for all material categories but wood (Fig.~\ref{fig:top5}-a), with an average overlap of 3.2 of 5. Based on these results, we consider C-MLP as the best performing model.

\textbf{Samples proximity analysis} -- 
When visualizing the C-MLP model in a t-SNE plot (Fig.~\ref{fig:main}-c), we observed a very similar pattern as in human ratings with woods and fabrics being clearly separated from other category clusters (see the supplementary material for a multidimensional scaling (MDS) visualization). In Fig.~\ref{fig:main}-b,c  we also plot similarity matrices computed from model predictions and from human ratings for 68 test samples for visual comparison.

\begin{figure}[!ht]
\includegraphics[width=\columnwidth]{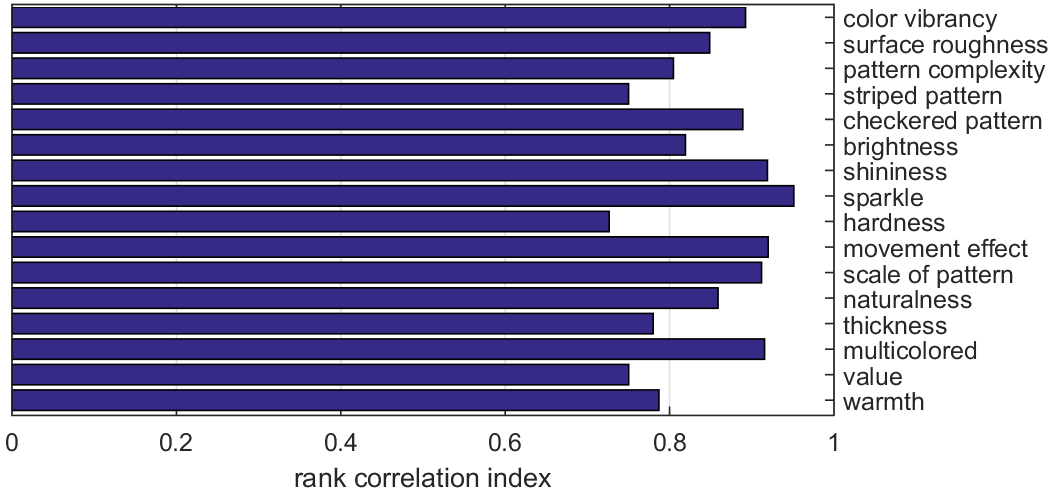}
\caption{\label{fig:rci}Spearman rank correlation index (RCI) for the order of the first 100 images when ranked by human ratings or model predictions, separately for each attribute. }
\end{figure}

\textbf{Ranking analysis} -- 
For each attribute, we also computed the Spearman rank correlation (RCI) to measure the overlap between the ordering of the first 100 images according to the human ratings versus the model predictions (Fig.~\ref{fig:rci}). The largest overlap (0.9$<$RCI$<$ 1.0) was obtained for ‘optical’ attributes like \emph{shininess}, \emph{sparkle}, \emph{movement effect}, \emph{scale of pattern}, and \emph{multicolored}. In contrast, attributes related to physical or more abstract material characteristics obtained lower scores (0.7$<$RCI$<$0.8), such as  \emph{hardness}, \emph{value}, \emph{thickness}, \emph{warmth}, and \emph{striped pattern}.

\begin{figure}[!ht]
\includegraphics[width=\columnwidth]{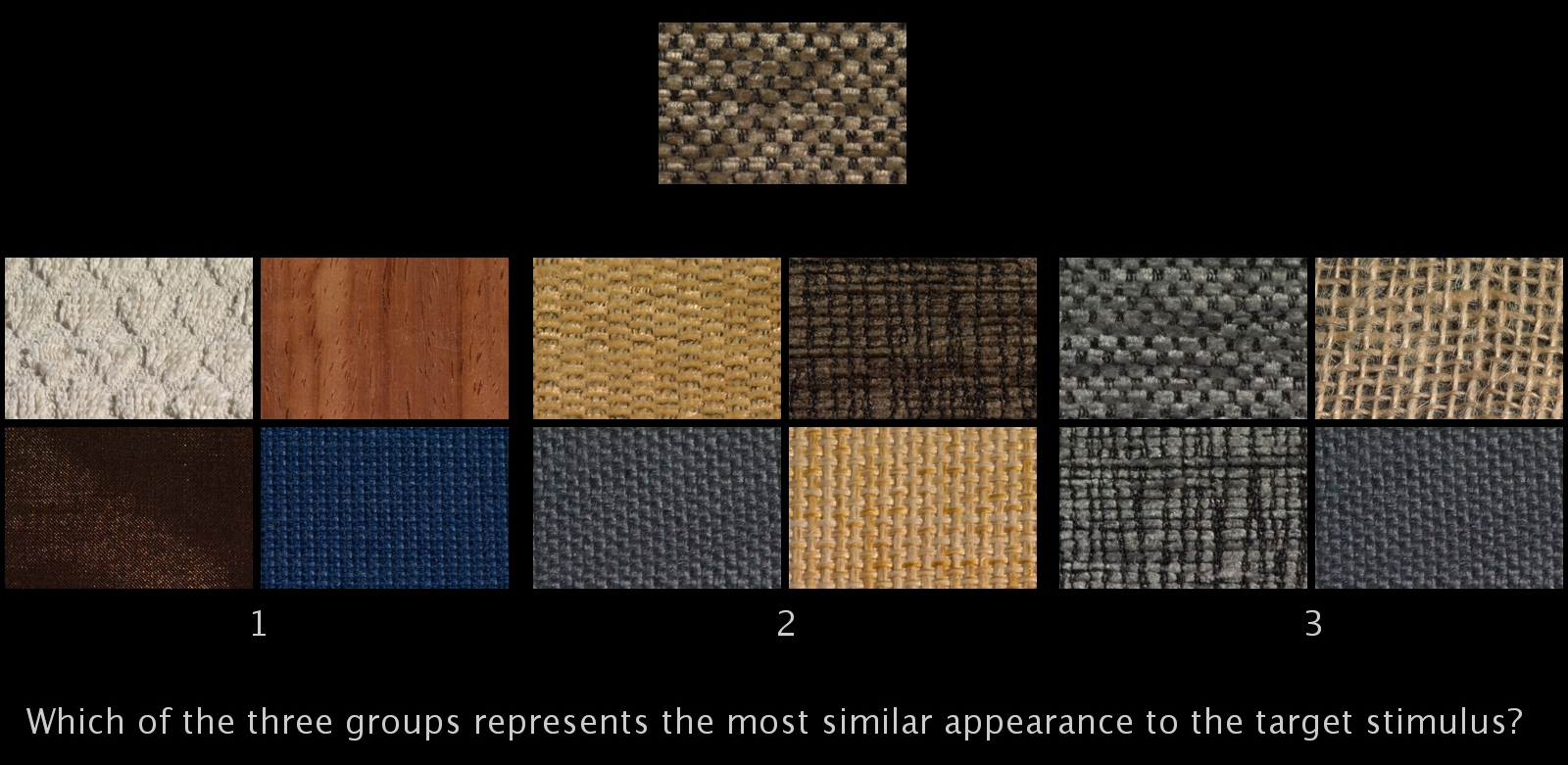}
\caption{\label{fig:validation}An example of one video frame from a trial of the validation experiment.  }
\end{figure}

\textbf{Study 5 -- Validation experiment} --To validate our C-MLP model in terms of predicting valid perceptual judgments, we presented each material test sample together with three groups of four samples (Fig.~\ref{fig:validation}). One group contained the four closest material samples selected based on human ratings, one group those selected based on model predictions, and one group four randomly selected samples. 22 online observers participated in the study and were asked: \emph{Which of the three groups represents the most similar appearance to the target stimulus?}. The order of groups on the screen was random.

The results show that, on average 41\% of observers preferred the group representing the human ratings, while 49\% preferred that based on the winning C-MLP model (Fig.~\ref{fig:top5}-b).

\subsection{Material Retrieval}

\begin{figure*}
\includegraphics[width=\textwidth]{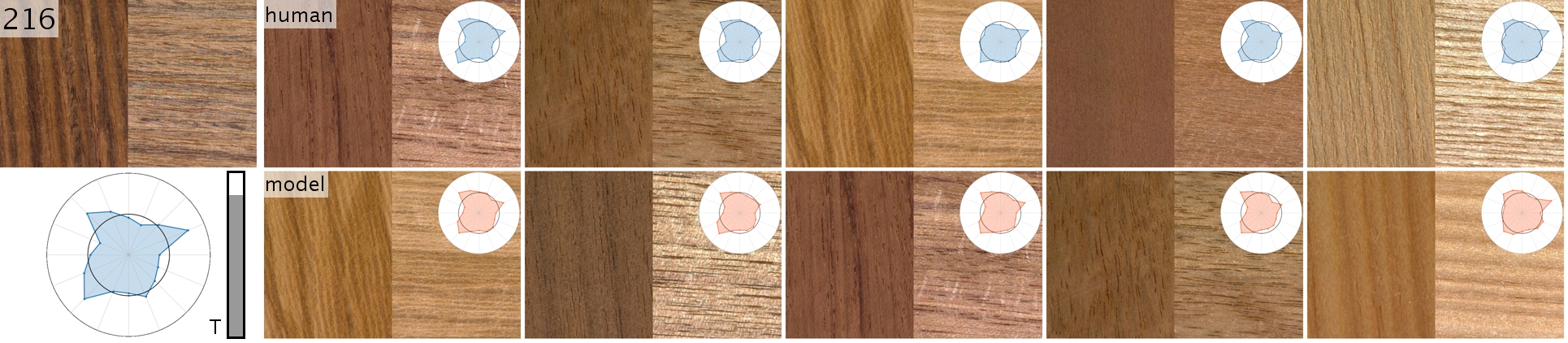}
\includegraphics[width=\textwidth]{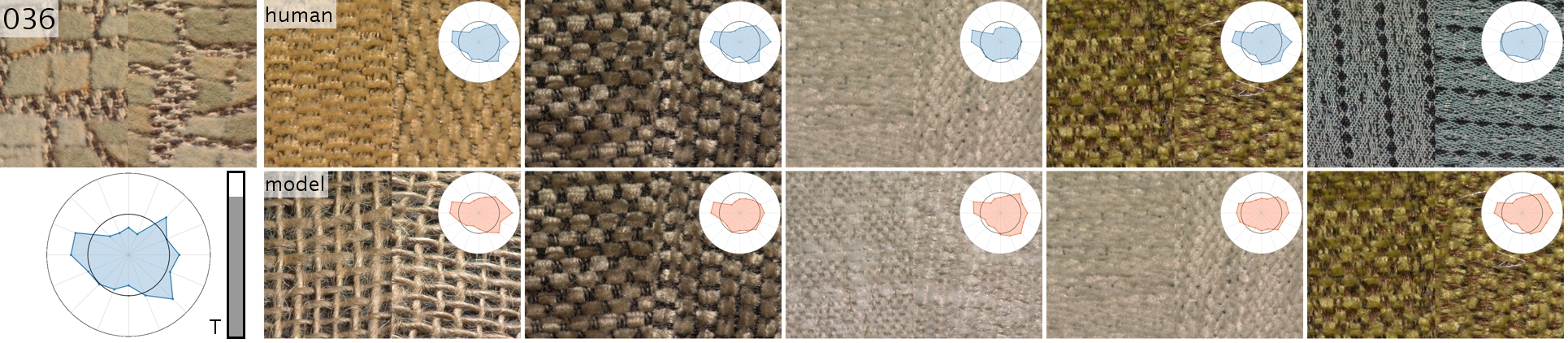}
\includegraphics[width=\textwidth]{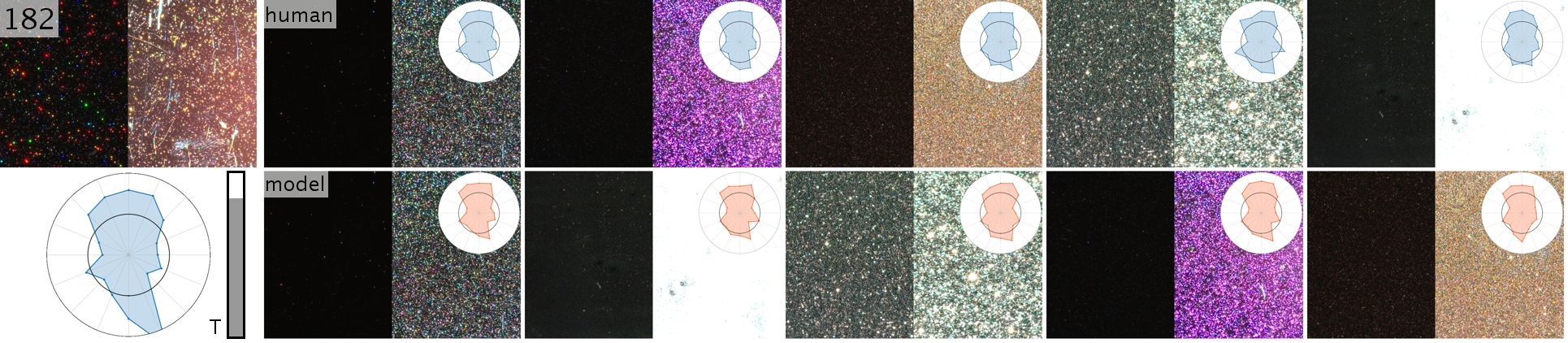}
\includegraphics[width=\textwidth]{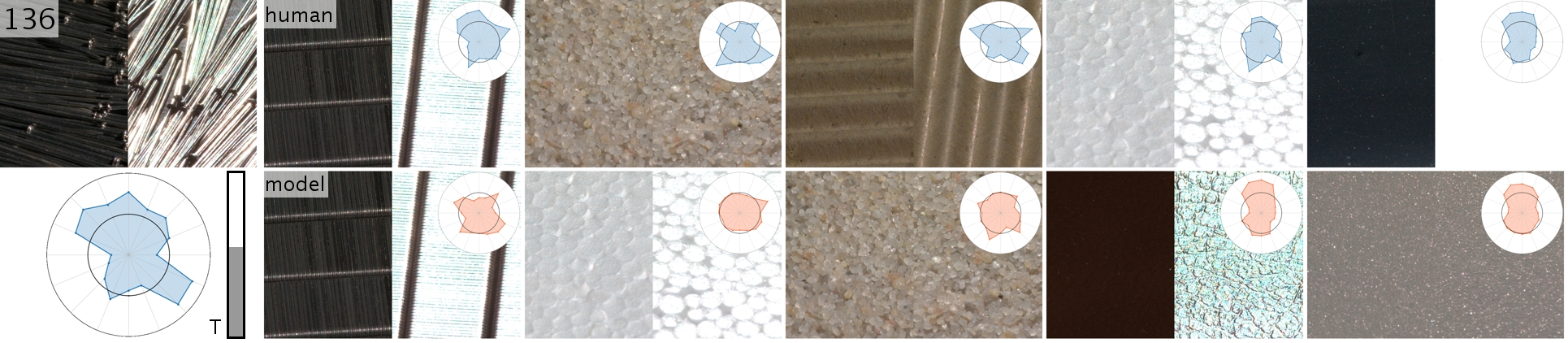}
\includegraphics[width=\textwidth]{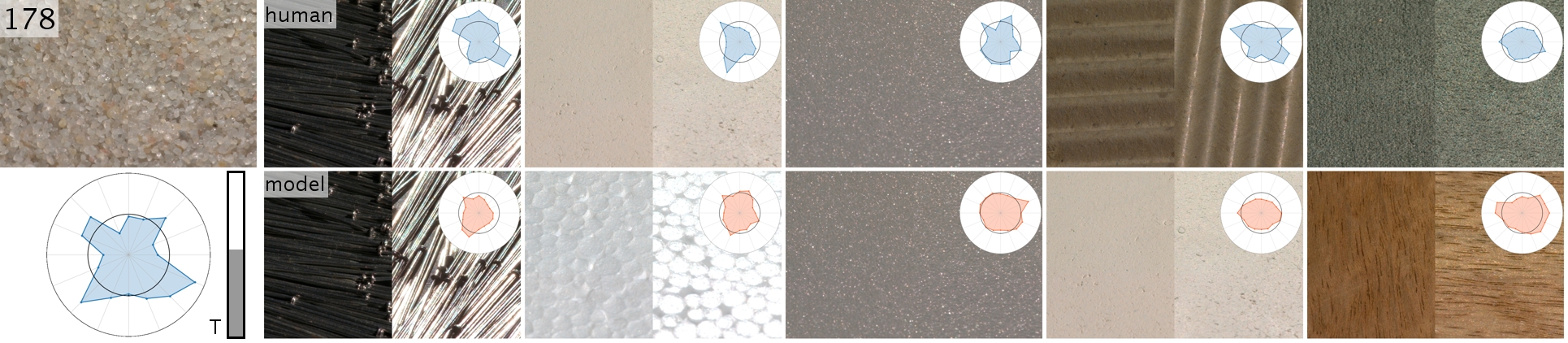}
\vspace{-0.2cm}
\caption{\label{fig:search}Example test material samples (left) with their visual fingerprints (below), followed by the five most similar samples according to human ratings (blue fingerprints) and model predictions (red fingerprints), identified by the strongest correlation coefficients between attributes. We show results for material samples of different typicality, indicated by the grey bar plot. All samples are depicted by one half of a specular and non-specular frame.}
\vspace{-0.3cm}
\end{figure*}

In the following, we can also use the distances between visual fingerprints (\ref{eq:1}) to identify materials with similar appearance. In Fig.~\ref{fig:search}, we plot for exemplary test material samples (left) the five most similar samples according to both human ratings (blue fingerprints) and model predictions (red fingerprints). The results suggest that the retrieval is highly effective when samples of similar appearance are available in the dataset, for example, for wood (216), fabric (036), or coating (182). However, even for more sparse categories, the retrieval identifies plausible samples, for example, for pins (136) or sand (178). We can quantify the similarity to other samples or ‘typicality’ of each sample by averaging across the distances to the 10\% most similar samples (Fig.~\ref{fig:search}, grey bar plots). The retrieval results for all materials are provided in the supplementary material.

\subsection{Material Retrieval in the Wild using Smartphone Shots}

\begin{figure*}[!ht]
\includegraphics[width=0.99\textwidth]{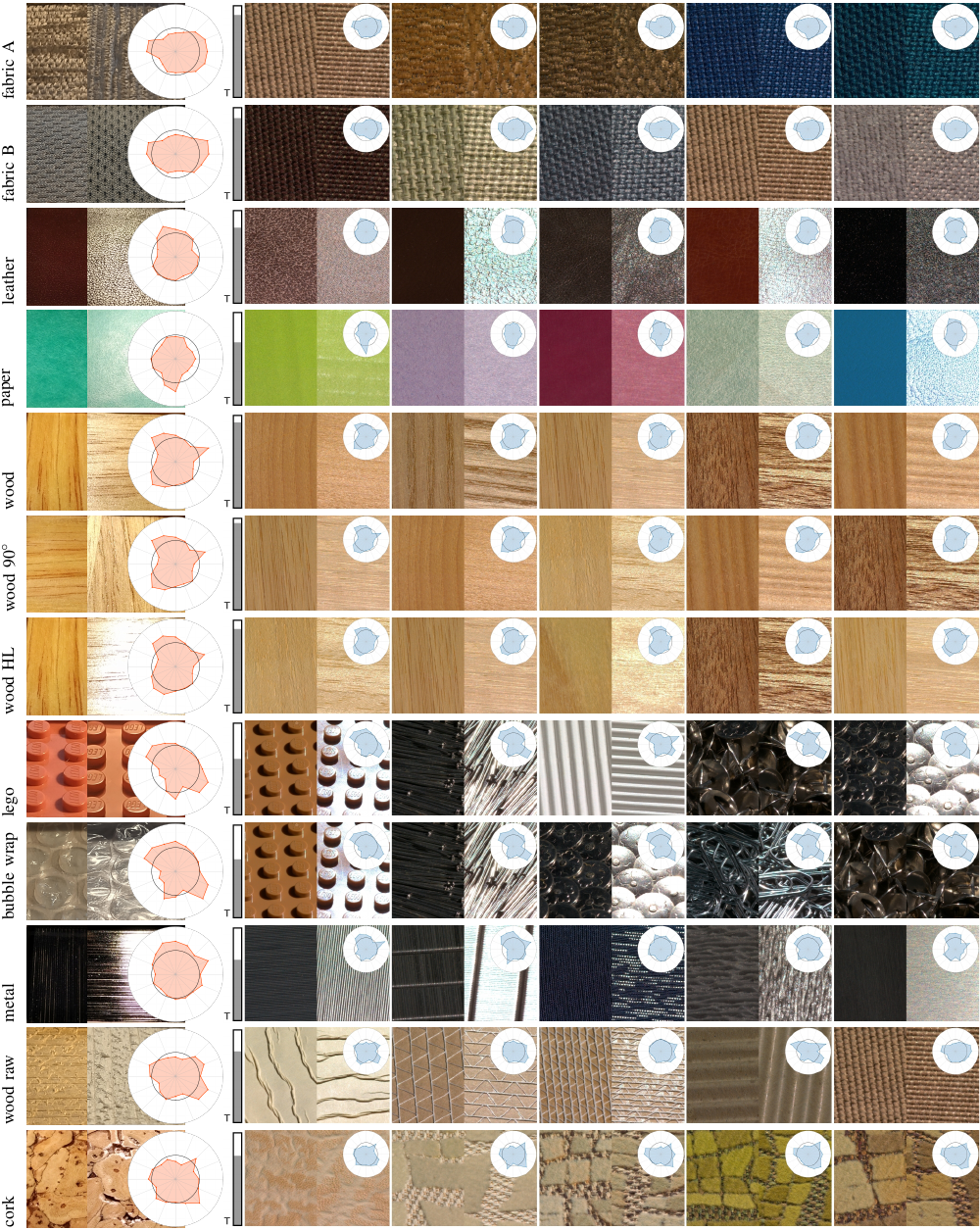}
\caption{\label{fig:mobile}Material retrieval of the five closest material samples from our dataset  for two smartphone close-up shots (non-specular, near-specular) of materials in the wild. A typicality of the material samples is indicated by the grey bar plot.}
\end{figure*}

To test our material retrieval method in the wild, we took smartphone close-up shots of different material surfaces, each under non-specular and near-specular conditions (light opposite the camera or on the left side of the sample). Then we cropped the images to obtain an area of interest 26$\times$26 mm, rescaled to 512$\times$512 pixels, and used our C-MLP model to predict the visual fingerprints. We used standard lens settings and a capturing distance of ~30 degrees. The captured images are not corrected for perspective, white balance or color.  Fig.~\ref{fig:mobile} dataset (grey bars) and the five most similar material samples from the set. We obtained very good matches from our dataset for standard materials like fabric, leather, paper, wood. To measure the effect of orientation, we included rotated shots of the same wood material, resulting in very similar predicted fingerprints and retrieval. We also tested less typical materials resulting in retrieval of material samples with similar visual features. More examples are provided in the supplementary material.

\section{Discussion}

In this paper, we introduce (1) 16 intuitive perceptual material appearance attributes that can be used for effective material identification and retrieval, and (2) a computational model for their prediction directly from material image data. 

\subsection{Advantages of Visual Fingerprinting}

Our method allows material retrieval according to perceptual appearance attributes (visual fingerprint) which therefore align with human observers’ intuitions about material similarity. The model is trained on $>$100,000 human rating responses, and can estimate human appearance judgements of any novel material from image data. As a result of capturing materials for both non-specular and specular behavior, our experiments also suggest some degree of invariance to orientation, the exact illumination and viewing angles. 


\subsection{Computational and Memory Demands}

We compared the performance of all inference models with respect to their complexity, by computing the Akaike information criterion AIC for the for 68 test samples (Tab.~\ref{tab:speed}, first column) . The results show \cite{burnham04multimodel} the lowest complexity for S-MLP and C-2NN as they rely on very small sets of parameters when compared to the texture synthesis T-MLP and CLIP-based C-MLP models that both have a more complex MLP structure.
The compared methods also vary in terms of model loading or initialization overhead, inference speed and memory footprint (Tab.~\ref{tab:speed}). The model based on statistics (S-MLP) requires precomputing some data structures for faster inference while the CLIP-based models require time-demanding loading of models to memory and have larger memory demands. Our results show that after initialization the CLIP features allow the highest inference speed.

\begin{table}
\caption{\label{tab:speed}A comparison of the tested models in terms of the Akaike information criterion (AIC) and the speed of the visual fingerprint inference from material images (including the one-time model loading) and model memory footprints. All tests were done on an Intel Core i7 with 16 GB RAM.}
\begin{tabular}{|l|r|rrr|}
\hline
  method & AIC & Initialization     & Inference & Model \\
             &       & overheads [s] & speed [s] & size [MB]\\
  \hline
  \textbf{T-MLP} (CPU)        & 667.6          &\textbf{0.0}  & 0.62 & 1.5 \\
  \textbf{S-MLP} (CPU)        & 2.4             &1.6  & 0.4 & \textbf{0.004} \\
  \textbf{C-2NN} (CPU)        & \textbf{1.6} &1.7  & 0.07 & 337.1 \\
  \textbf{C-MLP} (CPU/GPU) & 666.9          &1.7  & \textbf{0.07}/0.04 & 340.1 \\
  \hline
\end{tabular}
\end{table}

\subsection{Limitations and Future Directions}

\textit{Number of samples} --
Our dataset of 347 materials was carefully selected from a portfolio of real-world materials from different categories, and is one of the largest sets used in a psychophysical analysis to date. However, the number of samples per category varies, with the most samples for fabric and wood. This might have biased the model results, which is also suggested by the clustering in Fig.~\ref{fig:main}.

\textit{Observation and illumination geometry} --
 In contrast to previous studies we used video stimuli, providing observers with rich information on the dynamic appearance of materials under different viewing angles, including both specular and non-specular conditions. However, this is of course still a limited subset of all possible lighting-sample-viewer configurations. For example, we had to limit the camera and light trajectories to produce videos of reasonable length. Therefore, we did potentially not account for all specific retroreflective, goniochromatic, or anisotropic behaviors that for some materials might result from further changes in viewing and lighting polar angles.

\textit{Sensitivity to exact geometries} --  Capturing exact geometries is not always possible, however, the C-MLP model is quite robust for differences up to 15$^\circ$ as demonstrated on a comparison of images obtained by smartphone camera and comparisons between video frames and BTF data (see supplementary). On the other hand, using the same geometries will produce more consistent results as any angular deviation can lead to slightly different fingerprints and  material retrieval results (cf. \emph{wood} vs. \emph{wood HL} in Fig.~\ref{fig:mobile}). Our experiments demonstrate consistent performance when the camera view remains fixed while the light source moves, making mobile capture more convenient (see supplementary material for details).
 
\textit{Fixed sample scale} -- 
Although our experiments demonstrate a certain robustness to material scale variations, our current C-MLP model is limited to a fixed resolution and a fixed level of detail of the material surface (sample area of 26$\times$26 mm, resampled to a resolution of 512$\times$512 pixels). Also, our analysis is limited to materials with relatively fine and stationary textures and cannot describe visual behavior beyond our sample size, i.e., textures with too low spatial frequencies or too slow gradient changes across the sample. 

\textit{Open questions and future directions} --
While we tested statistical and deep learning features that proved to predict the human visual encoding of image data well \cite{kaniuth24high} and \cite{vacher21the}, there are of course other features which might potentially perform even better.
However, we consider this work as a proof-of-concept study, which can be extended in the future by testing different predictive models. To support this endeavor, we have made all stimuli (video and image data), rating responses, code for statistical analyses, and the trained C-MLP model available via a public repository. Finally, we provide an online application that evaluates the visual fingerprints of novel materials on demand.

\section{Conclusions}
\label{sec:conc}

In a series of psychophysical studies involving 347 material samples across various categories, we identified a set of 16 material attributes and obtained $>$100,000 observer ratings that define a visual fingerprint for each sample. 
We tested various image features and inference methods and identified a multi-layer perceptron model using deep learning features from a CLIP model as the best performing model to identify a mapping between material image statistics and human ratings.
Our findings suggest that this model can be used to predict visual fingerprints of material samples that support intuitive comparisons between materials and the retrieval of materials of similar appearance from the dataset. We validated the method in the wild by obtaining visual fingerprints for real-world materials from close-up smartphone shots. For this, we also provide users with an online app for an intuitive interface between material images and their interpretable perceptual attributes that will allow an efficient organization of materials in many applications.

\bibliographystyle{alpha}
\bibliography{references}

\newcommand{\etalchar}[1]{$^{#1}$}
\begin{thebibliography}{DGVG{\etalchar{+}}23}

\bibitem[BA04]{burnham04multimodel}
Kenneth~P Burnham and David~R Anderson.
\newblock Multimodel inference: understanding aic and bic in model selection.
\newblock {\em Sociological methods \& research}, 33(2):261--304, 2004.

\bibitem[Bro66]{brodatz66}
P.~Brodatz.
\newblock {\em A Photographic Album for Artists and Designers (Brodatz Texture
  Database)}.
\newblock Dover Publications, 1966.

\bibitem[DGVG{\etalchar{+}}23]{deschaintre23visual}
Valentin Deschaintre, Julia Guerrero-Viu, Diego Gutierrez, Tamy Boubekeur, and
  Belen Masia.
\newblock The visual language of fabrics.
\newblock {\em ACM Trans. Graph.}, 42(4), jul 2023.

\bibitem[DvGNK99]{dana99reflectance}
K.J. Dana, B.~van Ginneken, S.K. Nayar, and J.J. Koenderink.
\newblock Reflectance and texture of real-world surfaces.
\newblock {\em ACM Trans. on Graphics}, 18(1):1--34, 1999.

\bibitem[FDA03]{fleming03realworld}
R.~W. Fleming, R.~O. Dror, and E.~H. Adelson.
\newblock Real-world illumination and perception of surface reflectance
  properties.
\newblock In {\em Journal of Vision}, volume~3, pages 347--368, 2003.

\bibitem[FDL{\etalchar{+}}24]{filip24comprehensive}
J.~Filip, F.~Dechterenko, J.~Lukavsky, R.~Fleming, and F.~Schmidt.
\newblock Comprehensive perceptual analysis and rating of material properties
  from video data.
\newblock In {\em MCMI workshop (ICPR 2024), to appear in LNCS, Springer},
  2024.

\bibitem[FLD{\etalchar{+}}24]{filip24perceptual}
J.~Filip, J.~Lukavsk{\'y}, F.~D{\v e}cht{\v e}renko, F.~Schmidt, and R.~W.
  Fleming.
\newblock Perceptual dimensions of wood materials.
\newblock {\em Journal of Vision}, 5:12--12, 5 2024.

\bibitem[Fle14]{fleming14visual}
Roland~W. Fleming.
\newblock Visual perception of materials and their properties.
\newblock {\em Vision Research}, 94(0):62 -- 75, 2014.

\bibitem[FV14]{filip14template}
J.~Filip and R.~V{\'a}vra.
\newblock Template-based sampling of anisotropic {BRDF}s.
\newblock {\em Computer Graphics Forum}, 33(7):91--99, October 2014.

\bibitem[FV23]{filip23characterization}
J.~Filip and V.~Vilimovsk{\'a}.
\newblock Characterization of wood materials using perception-related image
  statistics.
\newblock {\em Journal of Imaging Science and Technology}, 5:1--9, 2023.

\bibitem[FVH{\etalchar{+}}13]{filip13brdf}
J.~Filip, R.~Vavra, M.~Haindl, P.~Zid, M.~Krupicka, and V.~Havran.
\newblock {BRDF} slices: Accurate adaptive anisotropic appearance acquisition.
\newblock In {\em Conference on Computer Vision and Pattern Recognition, CVPR},
  pages 4321--4326, 2013.

\bibitem[FWG13]{fleming13perceptual}
Roland~W Fleming, Christiane Wiebel, and Karl Gegenfurtner.
\newblock Perceptual qualities and material classes.
\newblock {\em Journal of vision}, 13(8):9--9, 2013.

\bibitem[GGG{\etalchar{+}}16]{guarnera16brdf}
D.~Guarnera, G.C. Guarnera, A.~Ghosh, C.~Denk, and M.~Glencross.
\newblock Brdf representation and acquisition.
\newblock {\em Computer Graphics Forum}, 35(2):625--650, 2016.

\bibitem[HLM07]{ho07conjoint}
Y.X. Ho, M.S. Landy, and L.T. Maloney.
\newblock Conjoint measurement of gloss and surface texture.
\newblock {\em Psychological Science}, 19:194--204, 2007.

\bibitem[JWD{\etalchar{+}}14]{jarabo14effects}
A.~Jarabo, H.~Wu, J.~Dorsey, H.~Rushmeier, and D.~Gutierrez.
\newblock Effects of approximate filtering on the appearance of bidirectional
  texture functions.
\newblock {\em IEEE Transactions on Visualization and Computer Graphics},
  20(6):880--892, June 2014.

\bibitem[KMPH24]{kaniuth24high}
Philipp Kaniuth, Florian~P Mahner, Jonas Perkuhn, and Martin~N Hebart.
\newblock A high-throughput approach for the efficient prediction of perceived
  similarity of natural objects.
\newblock {\em bioRxiv}, pages 2024--06, 2024.

\bibitem[LL02]{long02hybrid}
H.~Long and W.K. Leow.
\newblock A hybrid model for invariant and perceptual texture mapping.
\newblock In {\em Pattern Recognition, 2002. Proceedings. 16th International
  Conference on}, volume~1, pages 135--138. IEEE, 2002.

\bibitem[LMS{\etalchar{+}}19]{lagunas19similarity}
Manuel Lagunas, Sandra Malpica, Ana Serrano, Elena Garces, Diego Gutierrez, and
  Belen Masia.
\newblock A similarity measure for material appearance.
\newblock {\em ACM Transactions on Graphics (TOG)}, 38(4):1--12, 2019.

\bibitem[McC96]{mccanny96observation}
C.~S. McCanny.
\newblock Observation and measurement of the appearance of metallic materials.
  part. 1. macro appearance.
\newblock {\em COLOR research and applications}, 21(4):292--304, 1996.

\bibitem[MKK{\etalchar{+}}00]{mojsilovic00vocabulary}
A.~Mojsilovic, J.~Kovacevic, D.~Kall, R.J. Safranek, and S.~Kicha~Ganapathy.
\newblock The vocabulary and grammar of color patterns.
\newblock {\em Image Processing, IEEE Transactions on}, 9(3):417--431, 2000.

\bibitem[MP90]{malik90preattentive}
J.~Malik and P.~Perona.
\newblock Preattentive texture discrimination with early vision mechanisms.
\newblock {\em JOSA A}, 7(5):923--932, 1990.

\bibitem[MPBM03]{matusik03data}
Wojciech Matusik, Hanspeter Pfister, Matt Brand, and Leonard McMillan.
\newblock A data-driven reflectance model.
\newblock {\em ACM Trans. Graph.}, 22(3):759--769, jul 2003.

\bibitem[NRH{\etalchar{+}}77]{nicodemus77geometrical}
F.E. Nicodemus, J.C. Richmond, J.J. Hsia, I.W. Ginsburg, and T.~Limperis.
\newblock Geometrical considerations and nomenclature for reflectance.
\newblock {\em NBS Monograph 160}, pages 1--52, 1977.

\bibitem[PDG{\etalchar{+}}08]{padilla08perceived}
S.~Padilla, O.~Drbohlav, P.R. Green, A.~Spence, and M.J. Chantler.
\newblock Perceived roughness of 1/f{$\beta$} noise surfaces.
\newblock {\em Vision Research}, 48(17):1791 -- 1797, 2008.

\bibitem[PFG00]{pellacini00toward}
F.~Pellacini, J.A. Ferwerda, and D.P. Greenberg.
\newblock Toward a psychophysically-based light reflection model for image
  synthesis.
\newblock In {\em 27th International Conference on computer Graphics and
  Interactive Techniques}, pages 55--64, 2000.

\bibitem[PS00]{portilla00parametric}
Javier Portilla and Eero~P Simoncelli.
\newblock A parametric texture model based on joint statistics of complex
  wavelet coefficients.
\newblock {\em International journal of computer vision}, 40:49--70, 2000.

\bibitem[PSH07]{pont07texture}
S.C Pont, P.~Sen, and P.~Hanrahan.
\newblock $2\frac{1}{2}$d texture mapping: real-time perceptual surface
  roughening.
\newblock In {\em 4th Symphosium on Applied Perception in Graphics and
  Vizualization}, pages 69--72, 2007.

\bibitem[RFWB07]{ramanarayanan07visual}
G.~Ramanarayanan, J.~Ferwerda, B.~Walter, and K.~Bala.
\newblock Visual equivalence: towards a new standard for image fidelity.
\newblock {\em ACM Transactions on Graphics}, 26(3):76:1--76:10, 2007.

\bibitem[RKH{\etalchar{+}}21]{radford21learning}
Alec Radford, Jong~Wook Kim, Chris Hallacy, Aditya Ramesh, Gabriel Goh,
  Sandhini Agarwal, Girish Sastry, Amanda Askell, Pamela Mishkin, Jack Clark,
  et~al.
\newblock Learning transferable visual models from natural language
  supervision.
\newblock In {\em International conference on machine learning}, pages
  8748--8763. PMLR, 2021.

\bibitem[RRL96]{ravishankar96towards}
A.~Ravishankar~Rao and G.L. Lohse.
\newblock Towards a texture naming system: Identifying relevant dimensions of
  texture.
\newblock {\em Vision Research}, 36(11):1649--1669, 1996.

\bibitem[Rus14]{rushmeier14MAM}
Holly Rushmeier.
\newblock The {MAM}2014 sample set.
\newblock In {\em Proceedings of the Eurographics 2014 Workshop on Material
  Appearance Modeling: Issues and Acquisition}, MAM '14, pages 25--26, 2014.

\bibitem[SCW{\etalchar{+}}21]{serrano21effect}
Ana Serrano, Bin Chen, Chao Wang, Michal Piovar{\v{c}}i, Hans-Peter Seidel,
  Piotr Didyk, and Karol Myszkowski.
\newblock The effect of shape and illumination on material perception: model
  and applications.
\newblock {\em ACM Transactions on Graphics (TOG)}, 40(4):1--16, 2021.

\bibitem[SDO{\etalchar{+}}22]{sawayama19visual}
M~Sawayama, Y~Dobashi, M~Okabe, K~Hosokawa, T~Koumura, T~Saarela, M~Olkkonen,
  and S~Nishida.
\newblock Visual discrimination of optical material properties: a large-scale
  study.
\newblock {\em Journal of Vision}, 22(2):17, 2022.

\bibitem[SGM{\etalchar{+}}18]{serrano18intuitive}
Ana Serrano, Diego Gutierrez, Karol Myszkowski, Hans-Peter Seidel, and Belen
  Masia.
\newblock An intuitive control space for material appearance.
\newblock {\em ACM Transactions on Graphics (TOG)}, 35(6):1--12, 2018.

\bibitem[SL23]{subias2023wild}
J~Daniel Subias and Manuel Lagunas.
\newblock In-the-wild material appearance editing using perceptual attributes.
\newblock {\em Computer Graphics Forum}, 42(2):333--345, 2023.

\bibitem[SLRA13]{sharan13recognizing}
Lavanya Sharan, Ce~Liu, Ruth Rosenholtz, and Edward~H. Adelson.
\newblock Recognizing materials using perceptually inspired features.
\newblock {\em International Journal of Computer Vision}, 103(3):348--371, Jul
  2013.

\bibitem[SN13]{schwartz13visual}
G.~Schwartz and K.~Nishino.
\newblock Visual material traits: Recognizing per-pixel material context.
\newblock In {\em 2013 IEEE International Conference on Computer Vision
  Workshops}, pages 883--890, Dec 2013.

\bibitem[SN15]{schwartz13automatically}
G.~Schwartz and K.~Nishino.
\newblock Automatically discovering local visual material attributes.
\newblock In {\em 2015 IEEE Conference on Computer Vision and Pattern
  Recognition (CVPR)}, pages 3565--3573, June 2015.

\bibitem[SN19]{schwartz19recognizing}
Gabriel Schwartz and Ko~Nishino.
\newblock Recognizing material properties from images.
\newblock {\em IEEE transactions on pattern analysis and machine intelligence},
  42(8):1981--1995, 2019.

\bibitem[TH15]{tanaka15investigating}
Midori Tanaka and Takahiko Horiuchi.
\newblock Investigating perceptual qualities of static surface appearance using
  real materials and displayed images.
\newblock {\em Vision research}, 115:246--258, 2015.

\bibitem[TMY78]{tamura78textural}
H.~Tamura, S.~Mori, and T.~Yamawaki.
\newblock Textural features corresponding to visual perception.
\newblock {\em Systems, Man and Cybernetics, IEEE Transactions on},
  8(6):460--473, 1978.

\bibitem[tPP05a]{pas04comparison}
S.~F. te~Pas and S.~C. Pont.
\newblock A comparison of material and illumination discrimination performance
  for real rough, real smooth and computer generated smooth spheres.
\newblock In {\em 2nd Symp. on Applied Perception in Graphics and
  Visualization}, pages 57--58, 2005.

\bibitem[tPP05b]{pas05estimations}
S.~F. te~Pas and S.~C. Pont.
\newblock Estimations of light-source direction depend critically on material
  {BRDF}s.
\newblock {\em Perception, ECVP Abstract Supplement}, 34:212, 2005.

\bibitem[VB21]{vacher21the}
Jonathan Vacher and Thibaud Briand.
\newblock {The Portilla-Simoncelli Texture Model: towards Understanding the
  Early Visual Cortex}.
\newblock {\em {Image Processing On Line}}, 11:170--211, 2021.

\bibitem[VdMH08]{van08visualizing}
Laurens Van~der Maaten and Geoffrey Hinton.
\newblock Visualizing data using t-{SNE}.
\newblock {\em Journal of machine learning research}, 9(11):2579--2605, 2008.

\bibitem[VV97]{vanrell97afour}
M.~Vanrell and J.~Vitria.
\newblock A four-dimensional texture representation space.
\newblock In {\em Pattern Recognition and Image Analysis}, volume~1, pages
  245--250, 1997.

\bibitem[VVR97]{vanrell97multidimensional}
M.~Vanrell, J.~Vitria, and X.~Roca.
\newblock A multidimensional scaling approach to explore the behavior of a
  texture perception algorithm.
\newblock {\em Machine Vision and Applications}, 9(5/6):262--271, 1997.

\bibitem[WAKB09]{wills09toward}
Josh Wills, Sameer Agarwal, David Kriegman, and Serge Belongie.
\newblock Toward a perceptual space for gloss.
\newblock {\em ACM Trans. Graph.}, 28(4):103:1--103:15, September 2009.

\bibitem[WM01]{westlund01applying}
Harold~B. Westlund and Gary~W. Meyer.
\newblock Applying appearance standards to light reflection models.
\newblock In {\em Proceedings of the 28th annual conference on Computer
  graphics and interactive techniques}, SIGGRAPH '01, pages 501--51., New York,
  NY, USA, 2001. ACM.

\end{thebibliography}

\vfill
{\small
\begin{tabular}{p{1.5cm}p{6.5cm}}
\includegraphics[height=0.25\columnwidth]{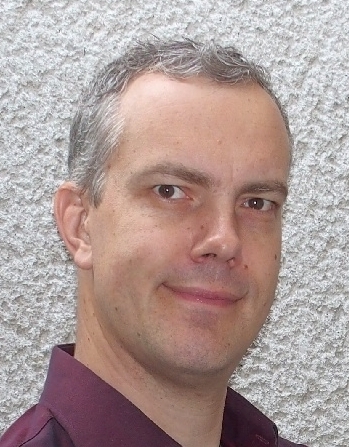}&
\vspace{-2.2cm}\textbf{Ji\v r\'\i~Filip} earned his PhD from the Czech Technical University in Prague in 2006. He did postdoc as a Marie-Curie fellow at Heriot-Watt University in Edinburgh. Since 2002, he has been a researcher at UTIA, The Czech Academy of Sciences. His work involves all aspects of material appearance, integrating methods from image processing, computer graphics, and visual psychophysics.\\
\includegraphics[height=0.25\columnwidth]{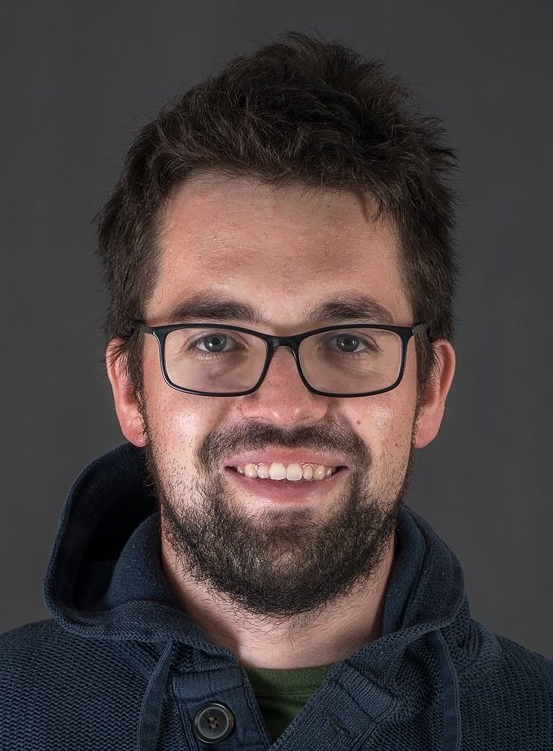}&
\vspace{-2.2cm}
\textbf{Filip D\v echt\v erenko} earned his PhD from Charles University in Prague, Czech Republic, in 2017. He is currently a researcher at the Institute of Psychology, part of the Czech Academy of Sciences. His research focuses on visual perception, attention, and memory.\\
\includegraphics[height=0.25\columnwidth]{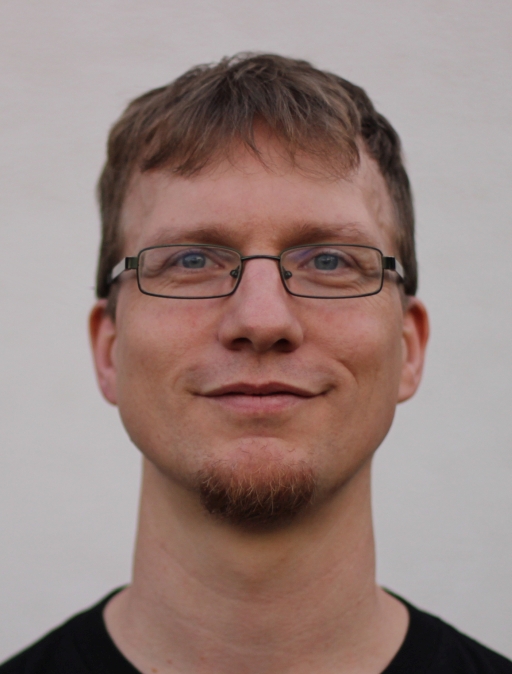}&
\vspace{-2.2cm}
\textbf{Filipp Schmidt} completed his PhD in Experimental Psychology at the University of Kaiserslautern, Germany, in 2014. 
Currently, he works as research coordinator in Giessen and is a principal investigator in the Collaborative Research Centre "Cardinal mechanisms of perception". His research interests include the perception of material and shape.\\
\includegraphics[height=0.25\columnwidth]{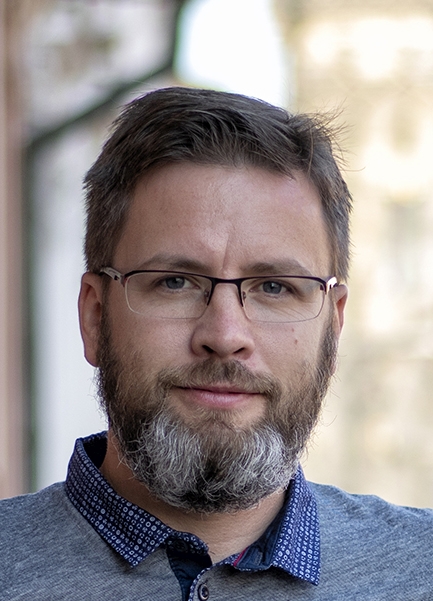}&
\vspace{-2.2cm}
\textbf{Ji\v r\'\i~Lukavský} received his PhD degree from Charles University, Prague, Czech Republic, in 2008. He currently works as a researcher in the Institute of Psychology, Czech Academy of Sciences. His research interests include visual perception, attention, and memory. \\
\includegraphics[height=0.25\columnwidth]{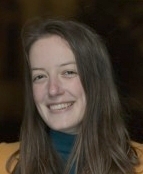}&
\vspace{-2.2cm}
\textbf{Veronika Vil\'\i movsk\'a} is a research assistant at UTIA, The Czech Academy of Sciences and a master student at Faculty of Information Technology, Czech Technical University in Prague. She is interested in data analysis and image processing. \\
\includegraphics[height=0.25\columnwidth]{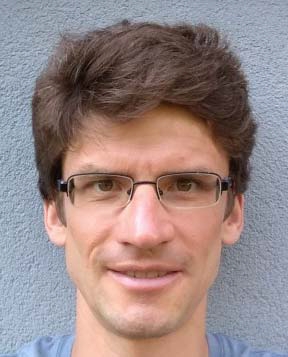}&
\vspace{-2.2cm}
\textbf{Jan Kotera} received the Ph.D. degree in mathematical modeling and computer science from Charles University, Prague. Since 2011, he has been with
UTIA, The Czech Academy of Sciences. His research interests include various aspects of digital image processing and computer vision, particularly blind deblurring and image restoration.\\
\includegraphics[height=0.25\columnwidth]{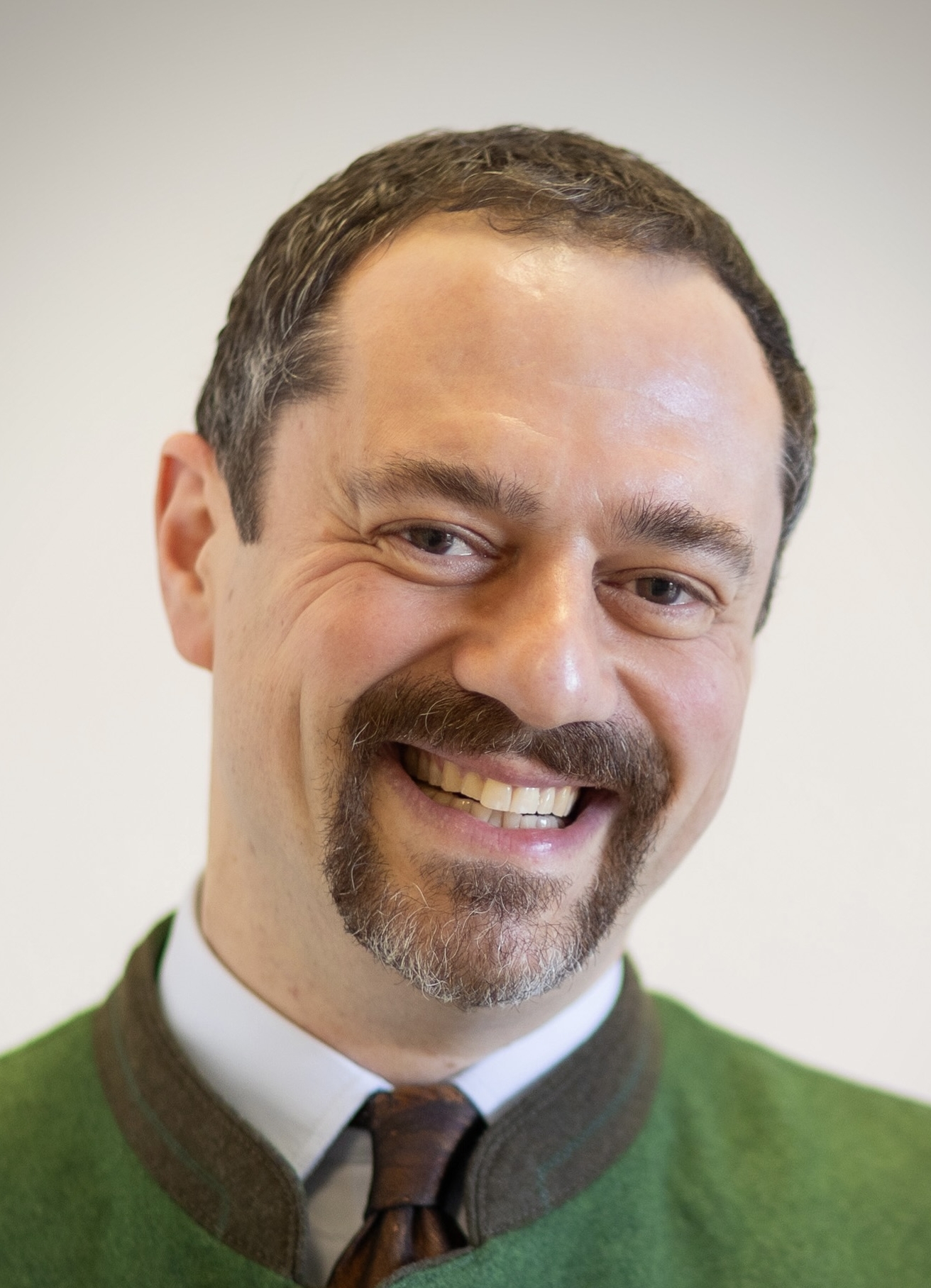}&
\vspace{-2.2cm}
\textbf{Roland Fleming} is the Kurt Koffka Professor of Experimental Psychology at Giessen University and the deputy Executive Director of the Center of Mind, Brain and Behavior. He studied at Oxford and MIT, and did a postdoc at the Max Planck Institute for Biological Cybernetics. He has been awarded the Vision Sciences Society Young Investigator Award as well as two ERC Grants. In 2022 he was elected Fellow of the Royal Society of Biology.\\
\end{tabular}
}

\end{document}